\documentclass[sn-mathphys,Numbered]{sn-jnl}

\usepackage{graphicx}%
\usepackage{multirow}%
\usepackage{amsmath,amssymb,amsfonts}%
\usepackage{amsthm}%
\usepackage{mathrsfs}%
\usepackage[title]{appendix}%
\usepackage{xcolor}%
\usepackage{textcomp}%
\usepackage{manyfoot}%
\usepackage{booktabs}%
\usepackage{algorithm}%
\usepackage{algorithmicx}%
\usepackage{algpseudocode}%
\usepackage{listings}%
\usepackage[T1]{fontenc}
\usepackage{upquote}

\usepackage{float}
\usepackage{titletoc}

\theoremstyle{thmstyleone}%

\theoremstyle{thmstyletwo}%

\theoremstyle{thmstylethree}%

\begin{document}

\title[Article Title]{CASCADE: Cumulative Agentic Skill Creation through Autonomous Development and Evolution}

\author[1,2]{\fnm{Xu} \sur{Huang}}\email{xu\_huang@berkeley.edu}

\author*[3,4]{\fnm{Junwu} \sur{Chen}}\email{junwu.chen@epfl.ch}

\author[1,2]{\fnm{Yuxing} \sur{Fei}}\email{yuxingfei@berkeley.edu}

\author[2]{\fnm{Zhuohan} \sur{Li}}\email{zhuohanli@lbl.gov}

\author*[3,4]{\fnm{Philippe} \sur{Schwaller}}\email{philippe.schwaller@epfl.ch}

\author*[1,2]{\fnm{Gerbrand} \sur{Ceder}}\email{gceder@berkeley.edu}

\affil[1]{Department of Materials Science and Engineering, University of California, Berkeley, California, United States}
\affil[2]{Materials Sciences Division, Lawrence Berkeley National Laboratory, California, United States}

\affil[3]{Laboratory of Artificial Chemical Intelligence (LIAC), 
  Institute of Chemical Sciences and Engineering, 
  \'{E}cole Polytechnique F\'{e}d\'{e}rale de Lausanne (EPFL), 
  Lausanne, Switzerland}
\affil[4]{National Centre of Competence in Research (NCCR) Catalysis, 
  \'{E}cole Polytechnique F\'{e}d\'{e}rale de Lausanne (EPFL), 
  Lausanne, Switzerland}

\abstract{
Large language model (LLM) agents currently depend on predefined tools or early-stage tool generation, limiting their adaptability and scalability to complex scientific tasks. We introduce CASCADE, a self-evolving agentic framework representing an early instantiation of the transition from ``LLM + tool use'' to ``LLM + skill acquisition''. CASCADE enables agents to master complex external tools and codify knowledge through two meta-skills: continuous learning via web search, code extraction, and memory utilization; self-reflection via introspection, knowledge graph exploration, and others. We evaluate CASCADE on SciSkillBench, a benchmark of 116 materials science and chemistry research tasks. CASCADE achieves a 93.3\% success rate using GPT-5, compared to 35.4\% without evolution mechanisms. We further demonstrate real-world applications in computational analysis, autonomous laboratory experiments, and selective reproduction of published papers. Along with human-agent collaboration and memory consolidation, CASCADE accumulates executable skills that can be shared across agents and scientists, moving toward scalable AI-assisted scientific research.
}

\maketitle

\section{Introduction}\label{sec1}

Tool use is not a unique ability of humans; what distinguishes humans is the cumulative acquisition of skills \cite{goodall1964tool, pargeter2019understanding, stout2017evolutionary}. Human technical culture advanced from opportunistic use of natural objects to the creation and mastery of sophisticated inventions, a transition from tool use to skill acquisition. This shift enabled specialized expertise to accumulate and transmit across generations, sustaining division of labor and cooperation at scale \cite{migliano2022origins, tomasello2019becoming}. We argue that large language model (LLM) agents now stand at an analogous inflection point. Embracing this progression could offer a path toward self-evolving and adaptive agents \cite{gao2025survey, jiang2025adaptation}, thereby supporting truly autonomous and scalable agentic intelligence.

Under the prevailing “LLM + tool use” paradigm, many agents rely on custom-built wrappers and predefined action spaces envisioned by humans \cite{schick2023toolformer, qin2024toolllm}. This design has enabled impressive progress across a range of applications, yet it also highlights an opportunity to further expand agents’ adaptability to increasingly complex tasks, where solving problems may require tools beyond those predefined at design time. Scalability represents a related challenge, as current agentic systems often depend on domain experts to manually curate detailed tool catalogs and task-specific prompts (Fig.~\ref{architecture}a, left). Recent efforts have begun to explore autonomous tool generation; however, such tools are typically constrained to Python’s built-in libraries and are often limited in scope \cite{cai2024large, yuan2024craft}. Moreover, while agents can increasingly interact with external APIs, sometimes augmented by web search or preselected resources, mechanisms for sustained skill mastery, human–agent collaboration, and memory-based consolidation remain relatively underexplored \cite{chenteaching, qiu2025alita, haque2025advanced, wangopenhands, tang2025autoagent, zheng2025skillweaver}. As a result, current automated tool-use pipelines, while effective in specific settings, remain at an early stage in the development of more robust, accurate, and self-evolving agentic capabilities, leaving substantial room for progress toward sophisticated and reusable skills.

As LLMs have become increasingly powerful in their reasoning and tool use capabilities, an expanding number of agentic systems focused on scientific discovery have emerged \cite{m2024augmenting, boiko2023autonomous, ruan2024automatic, mcnaughton2024cactus, huang2025biomni, chai2025scimaster, swanson2025virtual, matinvent, Takahara2025, nduma2025crystalyse}. Most systems primarily excel at literature search, hypothesis generation, and data analysis \cite{skarlinski2024language, ghareeb2025robin, mitchener2025kosmos}, yet challenges remain in executing complex, long-horizon experiments, both computationally and physically. To function as true AI co-scientists, agents benefit from autonomous execution capabilities that close the loop, enabling self-validation and iteration of ideas through tool implementation. Existing efforts typically rely on predefined tools \cite{zou2025agente, chandrasekhar2025automating} or implement early-stage approaches to tool generation \cite{villaescusa2025denario, yao2025operationalizing, jin2025stella, liu2025mattools, wolflein2025llm, miao2025paper2agent, gao2025democratizing, orimo2025parc, zhou2025toward, du2025accelerating}, with related limitations described in the literature \cite{swanson2025virtual, haque2025advanced, chen2024scienceagentbench}.

Here, we introduce CASCADE, an agentic framework facilitating AI co-scientists to develop cumulative executable skills. This framework represents an early instantiation of the transition from the ``LLM + tool use'' to ``LLM + skill acquisition'' paradigm, integrating multiple self-evolution mechanisms. Unlike traditional agentic systems that rely on problem-specific tools and prompts, CASCADE employs general problem-solving methodologies and mindsets to cultivate two meta-skills, continuous learning and self-reflection. Through natural language interfaces and multi-turn conversations, CASCADE lowers barriers to human-agent collaboration, allowing the agent to further evolve to solve more complex scientific problems. Additionally, it incorporates both session-wise memory, for referencing previous conversations, and consolidated memory, for retaining and retrieving key information, experience, and skill sets. These combined capabilities enable CASCADE to emulate human-like problem-solving behavior.

To evaluate CASCADE, we developed SciSkillBench, a comprehensive benchmark suite tailored for materials science and chemistry research. We conducted extensive baseline comparisons and ablation studies to validate CASCADE's capabilities. We also evaluated CASCADE on specific research tasks, including determining materials' piezoelectricity and analyzing systematic differences in predictions from machine learning interatomic potentials (MLIPs) \cite{deringer2019machine} trained on datasets generated at different density functional theory (DFT) levels \cite{geerlings2003conceptual}. Furthermore, we integrated CASCADE into an autonomous lab setting for real-world materials synthesis, characterization, and property measurement scenarios. CASCADE also successfully reproduced published results on Li intercalation voltages in rechargeable battery cathode materials \cite{isaacs2020prediction}. These results demonstrate CASCADE's ability to master complex external tools, codify knowledge and skills, and build sophisticated, reusable routines for human scientists, other agents, and scientific platforms \cite{gao2025democratizing}. Notably, CASCADE’s predefined tools are entirely domain-agnostic, containing no materials science or chemistry specific components, and the system prompts do not instruct the use of any domain-specific tools. Through context engineering, this domain-agnostic design could enable rapid transfer to other fields requiring complex tool interaction, such as software engineering and biology \cite{hua2025context, zhang2025agentic}. 

\section{Results}\label{sec2}

\subsection{CASCADE architecture}
\begin{figure}[H]
\centering
\includegraphics[width=1.01\textwidth]{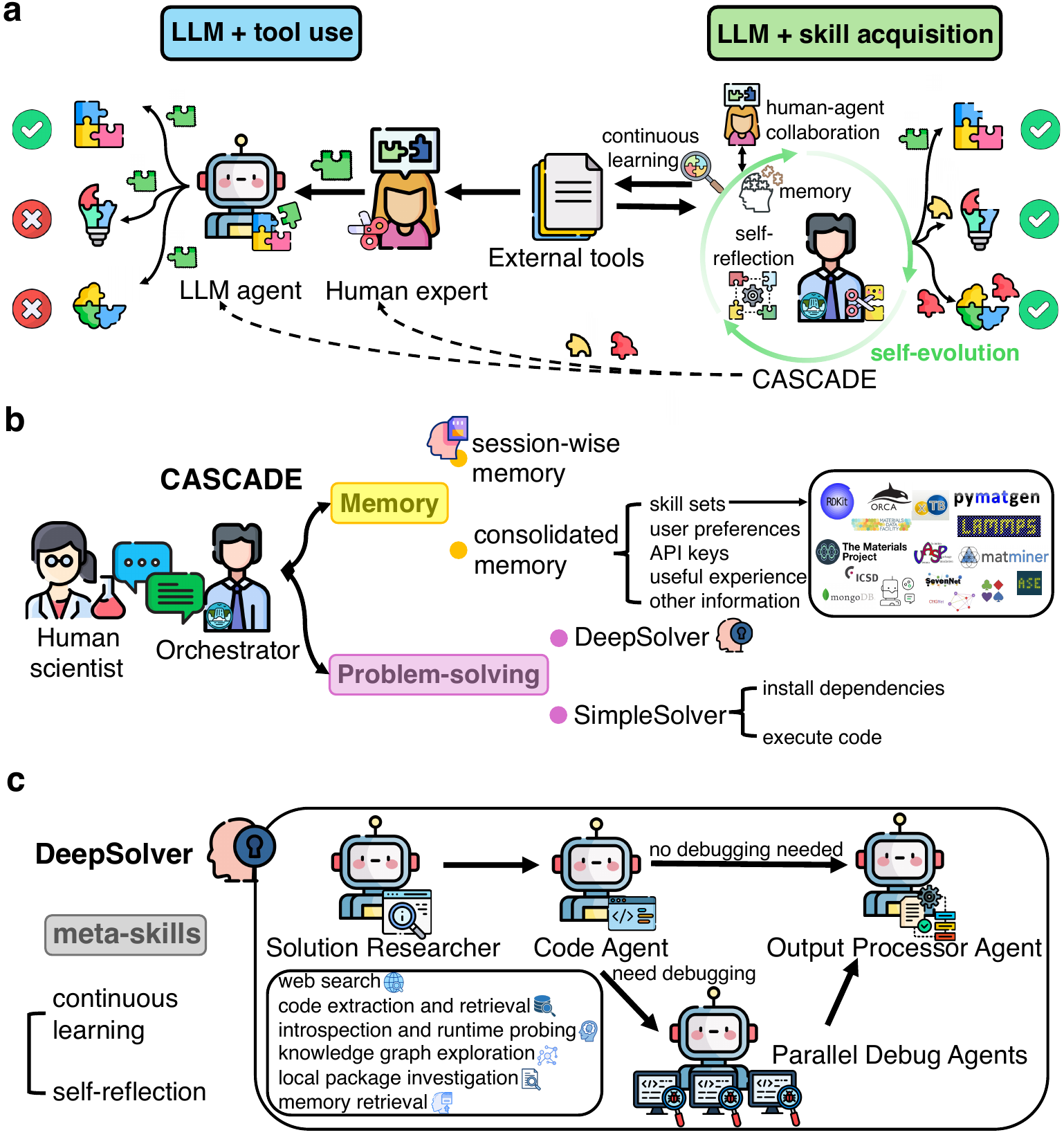}
\caption{\textbf{The ``LLM + skill acquisition'' paradigm and the CASCADE architecture.}
\textbf{a}, A puzzle-solving metaphor of the ``LLM + tool use'' versus the ``LLM + skill acquisition'' paradigm. On the left, agents rely on human experts to curate external tools. On the right, CASCADE showcases its ability to adeptly craft customized tools from complex external components, facilitating use by both human experts and LLM agents. 
\textbf{b}, The architecture of CASCADE. CASCADE facilitates multi-turn dialogues with human scientists through an interactive web interface, with persistent session management and consolidated memory in vector and graph databases. The Orchestrator agent within CASCADE selects between two solution pathways, the SimpleSolver or the DeepSolver, based on adaptable memory and task difficulty.
\textbf{c}, DeepSolver architecture. DeepSolver coordinates four specialized agents that collaboratively solve complex tasks while autonomously acquiring new tools and skills. It follows a sequential workflow: the Solution Researcher generates the initial code solution; the Code Agent executes the code; if debugging is required, the Debug Agents intervene; and finally, the Output Processor Agent processes the results.
}
\label{architecture}
\end{figure}

CASCADE is a self-evolving multi-agent system designed to function as an AI co-scientist by acquiring, refining and accumulating executable problem-solving skills over time (Fig.~\ref{architecture}a, right). As shown in Fig.~\ref{architecture}b, CASCADE comprises an Orchestrator agent that coordinates multi-turn dialogues with human scientists through an interactive web interface (\hyperref[methods]{Methods}). The system supports persistent session management, enabling users to initiate new conversations or resume previous sessions with full memory context. Upon receiving a user query, the Orchestrator retrieves relevant information from consolidated memory containing skill sets, user-specific preferences, API credentials, and accumulated experiences. The Orchestrator then selects between two problem-solving pathways. For straightforward queries or tasks solvable via adapted prior solutions, the Orchestrator uses the SimpleSolver for rapid response and resource efficiency (\hyperref[methods]{Methods}). SimpleSolver generates code, manages dependencies, and executes the code in an isolated environment. Successful executions return results directly; errors trigger escalation to DeepSolver. Complex problems bypass SimpleSolver entirely, proceeding directly to DeepSolver. 

DeepSolver utilizes a four-step sequential workflow with conditional parallel debugging (Fig.~\ref{architecture}c). Step 1: the Solution Researcher conducts web searches, extracts and retrieves code examples from URLs, and generates an initial code solution. Step 2: the Code Agent installs missing software dependencies, and executes the code, determining if debugging is needed. Step 3: If execution fails, three Debug Agent instances run concurrently, each employing different debugging strategies. Step 4: the Output Processor Agent evaluates all results, selects the best solution, and formats the output appropriately. Throughout this workflow, DeepSolver exhibits two meta-skills: continuous learning and self-reflection. Continuous learning enables the agent to acquire targeted external knowledge in real time through web search and code extraction, while adaptively leveraging previously accumulated skills across tasks through its memory system. Self-reflection operates beyond basic self-debugging, allowing the agent to evaluate solution quality, reason about failures and past experiences, and adapt its problem-solving strategies through diverse introspective and diagnostic tools, including code introspection, runtime probing, knowledge graph exploration, and local package investigation, among others (\hyperref[methods]{Methods}).

\subsection{Evaluating DeepSolver on scientific tasks}

DeepSolver is the key problem-solving engine that enables CASCADE to tackle real-world scientific research tasks. To rigorously assess its autonomous problem-solving capabilities, we evaluate DeepSolver with all memory tools disabled and without any human intervention. We create SciSkillBench, a diverse, carefully curated benchmark of 116 tasks for materials science and chemistry research. SciSkillBench also provides a reusable and automated evaluation suite for other agentic systems. 

\begin{figure}[!htbp]
\centering
\includegraphics[width=1\textwidth]{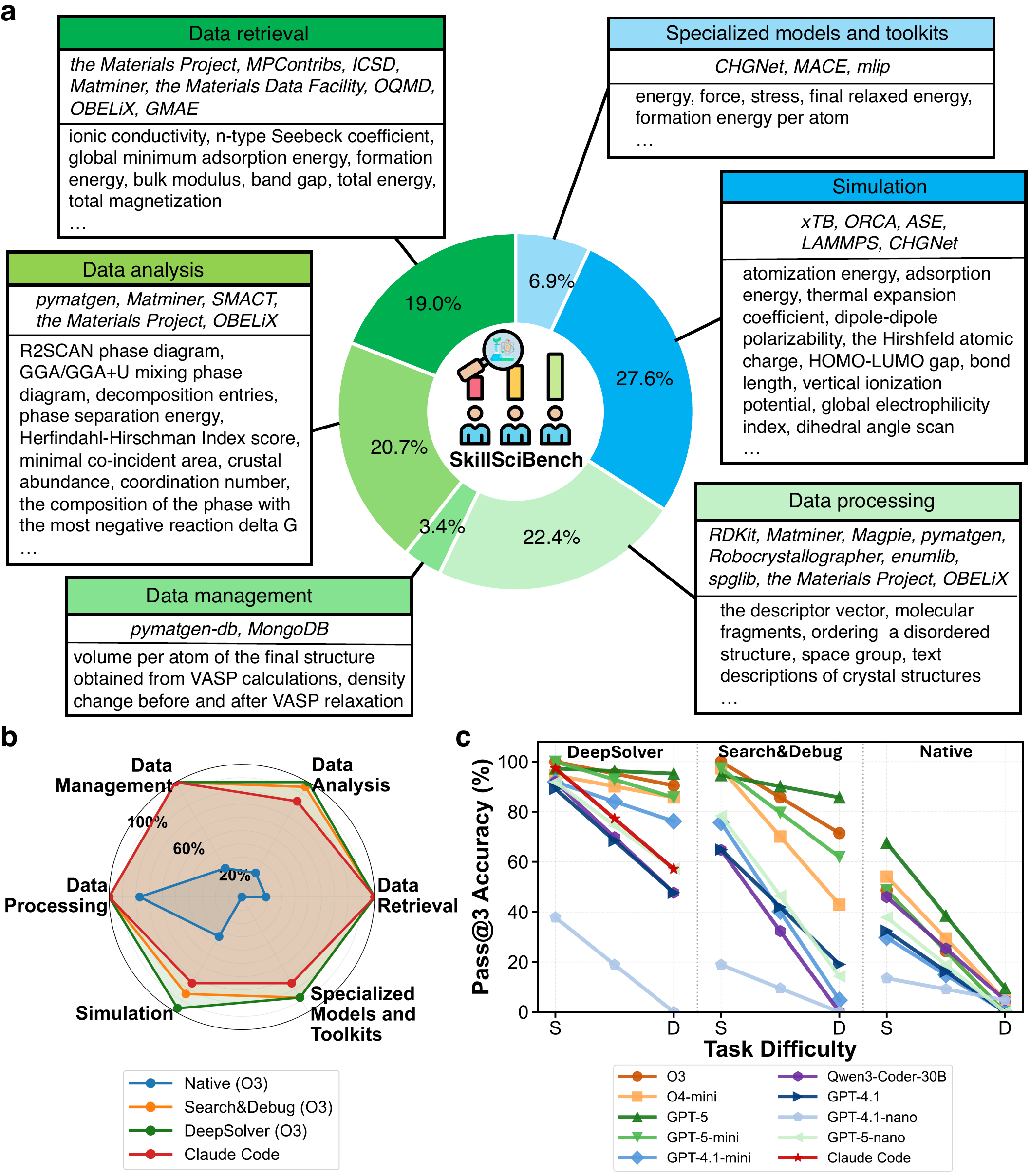}
\caption{\textbf{Task diversity in SciSkillBench and performance analysis.}
\textbf{a}, Overview of the diverse tasks in SciSkillBench, comprising six categories. The associated databases, packages, and software for each category are listed, alongside example quantities that the benchmarked system is required to obtain.
\textbf{b}, Pass@3 accuracy across all questions by task category for DeepSolver, Search\&Debug, Native (all using the OpenAI O3 model), and Claude Code. \textbf{c}, Pass@3 accuracy against task difficulty on Level 1 questions (S = Simple, D = Difficult). Each model uses a distinct color and marker. Claude Code (red star) appears in the DeepSolver section. 
}
\label{benchmark}
\end{figure}

As shown in Fig. \ref{benchmark}a, SciSkillBench encompasses two primary types of tasks: (i) 76 data-oriented tasks, which include 22 data-retrieval problems, 24 data-analysis problems, 4 data-management problems, and 26 data-processing problems, and (ii) 40 computation-oriented tasks, which consist of 32 simulation problems and 8 specialized model and toolkit-related problems. For each of the six categories, we highlight the key databases, packages, and software utilized, along with examples of specific quantities that the tested system is required to acquire (\hyperref[methods]{Methods}). The problems range from straightforward code adaptation and snippet stitching to undocumented queries that require strong scientific reasoning and exploratory coding, interaction with newly released packages or datasets absent from most model pre-training corpora, and navigation of outdated or misleading online documentation that can confuse agents. Many of these problems involve multiple external tools or different functions within a single tool, thus testing the agent's ability to effectively identify and chain these tools/functions to solve the problems. More detailed examples of the benchmark tasks can be found in \hyperref[si_tasks]{Appendix B.1}, where we provide some SciSkillBench instances as well as additional tasks not included in SciSkillBench.

Out of all tasks, 58 are categorized as Level 0, which reflect scenarios where users can specify key functions or core procedural components, but prefer not to handle low-level implementation details or error recovery. The remaining 58 tasks are categorized as Level 1, in which only high-level objectives are provided, with limited procedural guidance, requiring greater autonomy from the agent. Overall, Level 0 and Level 1 tasks differ in how user queries are formulated, with Level 0 queries more closely resembling those posed by computational scientists and Level 1 queries resembling those typically posed by experimental scientists.

We benchmark the performance of DeepSolver on SciSkillBench against three baselines (\hyperref[methods]{Methods}). The first baseline, referred to as Native, evaluates each LLM's inherent capabilities without any self-evolution. In this scenario, the Solution Researcher generates code once, the Code Agent executes it once, and the Output Processor Agent returns the result. The second baseline, designated as Search\&Debug Baseline (S\&D), reflects recent advancements in tool-making agents that enhance LLMs with capabilities such as web search and self-debugging, sometimes providing only one of these functions \cite{cai2024large, yuan2024craft, qiu2025alita, haque2025advanced, wangopenhands, tang2025autoagent, zheng2025skillweaver, villaescusa2025denario, yao2025operationalizing, jin2025stella, liu2025mattools, wolflein2025llm, miao2025paper2agent, gao2025democratizing, orimo2025parc, zhou2025toward}. In our implementation, we provide both search and debug functions to augment performance for this baseline. The Solution Researcher is enabled to conduct web search, while the Code Agent is required to engage in iterative self-debugging. For both of these baselines, as well as DeepSolver, we employ nine leading models as the LLMs for these multi-agent systems. These models include eight from OpenAI: GPT-5, GPT-5-mini, GPT-5-nano, O3, O4-mini, GPT-4.1, GPT-4.1-mini, and GPT-4.1-nano, in addition to one open-source model, Qwen3-Coder-30B-A3B-Instruct-FP8 (Qwen3-Coder-30B) \cite{yang2025qwen3}. For the third baseline, we implement a Claude Code agentic system leveraging Anthropic's Claude Agent software development kit (SDK), which invokes Claude Code CLI with all its built-in tools, including web search and self-debugging capabilities. We select Claude-Sonnet-4.5, Anthropic's highest-performing coding model at the time of testing, as the LLM for this system to probe the reach and limit of one of the strongest commercial coding agents on SciSkillBench.

\begin{table*}[!htbp]
    \centering
    \setlength{\tabcolsep}{2.5pt}
    \renewcommand{\arraystretch}{1.0}
    \resizebox{\textwidth}{!}{
      \begin{tabular}{l|l|ccc|ccc|ccc|ccc}
         \toprule
        \multicolumn{14}{c}{\textbf{Baseline Comparison (Native, Search\&Debug, DeepSolver)}} \\
        \midrule
        \multirow{2}{*}{Models} & \multirow{2}{*}{Metrics} & \multicolumn{3}{c|}{All Questions (\%)} & \multicolumn{3}{c|}{0-Level Questions (\%)} & \multicolumn{3}{c|}{1-Level Questions (\%)} & \multicolumn{3}{c}{Average Time (s)} \\
        \cmidrule(lr){3-5} \cmidrule(lr){6-8} \cmidrule(lr){9-11} \cmidrule(lr){12-14}
        & & Native & S\&D & DeepSolver & Native & S\&D & DeepSolver & Native & S\&D & DeepSolver & Native & S\&D & DeepSolver \\
        \midrule
        \multirow{4}{*}{GPT-5} & Success Rate & 35.36 & 89.74 & \textbf{93.26} & 39.53 & 94.71 & \textbf{96.47} & 31.21 & 84.80 & \textbf{90.06} & 240 & 504 & 588 \\
        \cmidrule(lr){2-14}
        & Pass@1 & 32.76 & 90.52 & \textbf{93.97} & 39.66 & 94.83 & \textbf{96.55} & 25.86 & 86.21 & \textbf{91.38} & & & \\
        & Pass@2 & 48.28 & 93.97 & \textbf{97.41} & 55.17 & 98.28 & \textbf{100.00} & 41.38 & 89.66 & \textbf{94.83} & & & \\
        & Pass@3 & 53.45 & 94.83 & \textbf{98.28} & 60.34 & 98.28 & \textbf{100.00} & 46.55 & 91.38 & \textbf{96.55} & & & \\
        \midrule
        \multirow{4}{*}{O3} & Success Rate & 23.05 & 80.76 & \textbf{91.84} & 26.59 & 85.88 & \textbf{97.70} & 19.54 & 75.72 & \textbf{85.80} & 153 & 352 & 407 \\
        \cmidrule(lr){2-14}
        & Pass@1 & 21.55 & 75.86 & \textbf{93.97} & 24.14 & 79.31 & \textbf{96.55} & 18.97 & 72.41 & \textbf{91.38} & & & \\
        & Pass@2 & 29.31 & 89.66 & \textbf{98.28} & 34.48 & 96.55 & \textbf{100.00} & 24.14 & 82.76 & \textbf{96.55} & & & \\
        & Pass@3 & 35.34 & 93.97 & \textbf{98.28} & 39.66 & 98.28 & \textbf{100.00} & 31.03 & 89.66 & \textbf{96.55} & & & \\
        \midrule
        \multirow{4}{*}{O4-mini} & Success Rate & 24.71 & 68.39 & \textbf{86.30} & 25.29 & 75.29 & \textbf{86.63} & 24.14 & 61.49 & \textbf{85.96} & 118 & 177 & 248 \\
        \cmidrule(lr){2-14}
        & Pass@1 & 24.14 & 62.93 & \textbf{82.61} & 24.14 & 67.24 & \textbf{84.48} & 24.14 & 58.62 & \textbf{80.70} & & & \\
        & Pass@2 & 34.48 & 80.17 & \textbf{91.30} & 36.21 & 86.21 & \textbf{91.38} & 32.76 & 74.14 & \textbf{91.23} & & & \\
        & Pass@3 & 37.93 & 84.48 & \textbf{94.78} & 39.66 & 91.38 & \textbf{96.55} & 36.21 & 77.59 & \textbf{92.98} & & & \\
        \midrule
        \multirow{4}{*}{GPT-5-mini} & Success Rate & 22.09 & 69.39 & \textbf{82.18} & 23.84 & 76.07 & \textbf{89.08} & 20.35 & 62.87 & \textbf{75.29} & 353 & 469 & 453 \\
        \cmidrule(lr){2-14}
        & Pass@1 & 18.97 & 71.55 & \textbf{83.62} & 20.69 & 82.76 & \textbf{91.38} & 17.24 & 60.34 & \textbf{75.86} & & & \\
        & Pass@2 & 30.17 & 85.34 & \textbf{93.97} & 29.31 & 91.38 & \textbf{100.00} & 31.03 & 79.31 & \textbf{87.93} & & & \\
        & Pass@3 & 34.48 & 88.79 & \textbf{97.41} & 37.93 & 93.10 & \textbf{100.00} & 31.03 & 84.48 & \textbf{94.83} & & & \\
        \midrule
        \multirow{4}{*}{GPT-4.1-mini} & Success Rate & 16.03 & 45.65 & \textbf{72.78} & 18.24 & 55.76 & \textbf{76.40} & 13.87 & 35.03 & \textbf{69.03} & 76 & 156 & 190 \\
        \cmidrule(lr){2-14}
        & Pass@1 & 15.52 & 42.61 & \textbf{70.69} & 15.52 & 51.72 & \textbf{74.14} & 15.52 & 33.33 & \textbf{67.24} & & & \\
        & Pass@2 & 21.55 & 59.13 & \textbf{85.34} & 24.14 & 70.69 & \textbf{89.66} & 18.97 & 47.37 & \textbf{81.03} & & & \\
        & Pass@3 & 23.28 & 64.35 & \textbf{89.66} & 27.59 & 77.59 & \textbf{93.10} & 18.97 & 50.88 & \textbf{86.21} & & & \\
        \midrule
        \multirow{4}{*}{\shortstack{Qwen3-Coder\\-30B-A3B\\-Instruct-FP8}} & Success Rate & 21.41 & 34.07 & \textbf{64.38} & 24.05 & 42.75 & \textbf{72.14} & 18.93 & 25.90 & \textbf{57.24} & 426 & 515 & 599 \\
        \cmidrule(lr){2-14}
        & Pass@1 & 20.87 & 35.29 & \textbf{65.22} & 22.81 & 41.18 & \textbf{70.69} & 18.97 & 29.41 & \textbf{59.65} & & & \\
        & Pass@2 & 30.43 & 48.04 & \textbf{76.52} & 35.09 & 56.86 & \textbf{81.03} & 25.86 & 39.22 & \textbf{71.93} & & & \\
        & Pass@3 & 37.39 & 52.94 & \textbf{80.00} & 43.86 & 62.75 & \textbf{82.76} & 31.03 & 43.14 & \textbf{77.19} & & & \\
        \midrule
        \multirow{4}{*}{GPT-4.1} & Success Rate & 19.83 & 40.46 & \textbf{62.82} & 22.99 & 47.67 & \textbf{71.68} & 16.67 & 33.33 & \textbf{54.02} & 67 & 85 & 187 \\
        \cmidrule(lr){2-14}
        & Pass@1 & 18.97 & 42.24 & \textbf{63.79} & 18.97 & 46.55 & \textbf{67.24} & 18.97 & 37.93 & \textbf{60.34} & & & \\
        & Pass@2 & 24.14 & 55.17 & \textbf{74.14} & 29.31 & 63.79 & \textbf{79.31} & 18.97 & 46.55 & \textbf{68.97} & & & \\
        & Pass@3 & 27.59 & 57.76 & \textbf{81.03} & 34.48 & 67.24 & \textbf{87.93} & 20.69 & 48.28 & \textbf{74.14} & & & \\
        \midrule
        \multirow{4}{*}{GPT-5-nano} & Success Rate & 16.95 & 38.79 & \textbf{57.27} & 17.82 & 40.80 & \textbf{60.82} & 16.09 & 36.78 & \textbf{53.76} & 146 & 196 & 394 \\
        \cmidrule(lr){2-14}
        & Pass@1 & 16.38 & 38.79 & \textbf{62.93} & 13.79 & 37.93 & \textbf{63.79} & 18.97 & 39.66 & \textbf{62.07} & & & \\
        & Pass@2 & 23.28 & 49.14 & \textbf{75.00} & 24.14 & 50.00 & \textbf{77.59} & 22.41 & 48.28 & \textbf{72.41} & & & \\
        & Pass@3 & 28.45 & 61.21 & \textbf{81.90} & 32.76 & 67.24 & \textbf{84.48} & 24.14 & 55.17 & \textbf{79.31} & & & \\
        \midrule
        \multirow{4}{*}{GPT-4.1-nano} & Success Rate & 6.38 & 5.71 & \textbf{19.92} & 7.56 & 3.57 & \textbf{24.60} & 5.20 & 7.88 & \textbf{15.38} & 52 & 63 & 178 \\
        \cmidrule(lr){2-14}
        & Pass@1 & 5.17 & 6.03 & \textbf{22.81} & 5.17 & 5.17 & \textbf{31.03} & 5.17 & 6.90 & \textbf{14.29} & & & \\
        & Pass@2 & 8.62 & 9.48 & \textbf{30.70} & 8.62 & 8.62 & \textbf{37.93} & 8.62 & 10.34 & \textbf{23.21} & & & \\
        & Pass@3 & 12.07 & 10.34 & \textbf{32.46} & 13.79 & 8.62 & \textbf{39.66} & 10.34 & 12.07 & \textbf{25.00} & & & \\
        \midrule
          \multicolumn{14}{c}{\textbf{Claude Code Baseline}} \\
          \midrule
          \multirow{2}{*}{Models} & \multirow{2}{*}{Metrics} & \multicolumn{3}{c|}{All Questions (\%)} & \multicolumn{3}{c|}{0-Level Questions (\%)} & \multicolumn{3}{c|}{1-Level Questions (\%)} & \multicolumn{3}{c}{Average Time (s)} \\
          \cmidrule(lr){3-5} \cmidrule(lr){6-8} \cmidrule(lr){9-11} \cmidrule(lr){12-14}
          & & \multicolumn{3}{c|}{Claude Code} & \multicolumn{3}{c|}{Claude Code} & \multicolumn{3}{c|}{Claude Code} & \multicolumn{3}{c}{Claude Code} \\
          \midrule
          \multirow{4}{*}{Claude-Sonnet-4.5} & Success Rate & \multicolumn{3}{c|}{82.47} & \multicolumn{3}{c|}{90.23} & \multicolumn{3}{c|}{74.71} & \multicolumn{3}{c}{239} \\
          \cmidrule(lr){2-14}
          & Pass@1 & \multicolumn{3}{c|}{82.76} & \multicolumn{3}{c|}{89.66} & \multicolumn{3}{c|}{75.86} & \multicolumn{3}{c}{} \\
          & Pass@2 & \multicolumn{3}{c|}{87.07} & \multicolumn{3}{c|}{91.38} & \multicolumn{3}{c|}{82.76} & \multicolumn{3}{c}{} \\
          & Pass@3 & \multicolumn{3}{c|}{87.93} & \multicolumn{3}{c|}{93.10} & \multicolumn{3}{c|}{82.76} & \multicolumn{3}{c}{} \\
          \bottomrule
      \end{tabular}
    }
\caption{\textbf{Baseline comparison results.} The upper section compares three systems (Native, Search\&Debug (S\&D), DeepSolver) across various models. The lower section shows Claude Code Baseline results using Claude-Sonnet-4.5. Success Rate represents the overall accuracy across all attempts, while Pass@k (k=1, 2, 3) indicates the proportion of questions with at least one correct answer among the first k attempts. Results are shown for all questions, as well as separately for 0-Level and 1-Level questions. Average time denotes the mean completion time per question in seconds. Bold values indicate the best performance for each model across the systems (Native, S\&D, DeepSolver).}
  \label{baseline}
  \end{table*}

For each of the above agentic systems, we conduct three independent repetitions for each benchmark question (\hyperref[methods]{Methods}). Table~\ref{baseline} presents the results for DeepSolver, Native, S\&D, and Claude Code. It includes the overall success rate, the average run time, success rates for Level 0 and Level 1 questions, and pass@k accuracy (k=1, 2, 3) for all questions, Level 0, and Level 1. Here, pass@k accuracy means an answer is a ``pass'' if at least one of the first k responses is correct. From Table~\ref{baseline}, we observe that CASCADE's DeepSolver subsystem performs best across all comparisons. For nearly all agentic systems, the success rate for Level 0 questions consistently exceeds that for Level 1 questions. Even the most advanced models struggle without evolutionary mechanisms (Native), with the best overall success rate at only 35\%. However, when models undergo effective evolution, particularly those models that can better understand our prompts and effectively utilize the general-purpose tools in DeepSolver, performance improvements are substantial. The highest overall success rate improvement is nearly 70\% over Native with the O3 model. For Level 0 tasks, the GPT-5, O3, and GPT-5-mini models achieve 100\% pass@2 accuracy using DeepSolver. This suggests potential for handling more complex Level 0 research tasks. Furthermore, DeepSolver shows considerable enhancements over S\&D across all models, with an average improvement of 17.53\% in overall success rate. Nevertheless, DeepSolver generally requires more time to output results compared to Native and S\&D. Notably, when paired with GPT-4.1-mini, DeepSolver achieves higher overall Pass@3 accuracy than Claude Code Baseline (89.66\% versus 87.93\%) with shorter execution time (190 s versus 239 s), despite GPT-4.1-mini having approximately 7.5 times lower input cost and 9.4 times lower output cost than Claude-Sonnet-4.5 at the time of testing. This demonstrates that our agentic framework can effectively leverage less capable yet more economical models to achieve competitive or superior performance.

To further investigate the performance advantages of DeepSolver’s self-evolution compared to the Native, S\&D, and Claude Code baselines, we assess each system’s efficacy across six task categories and two difficulty tiers (simple and difficult). In Fig.~\ref{benchmark}b, we present the pass@3 accuracy across all questions by task category for DeepSolver, S\&D, and Native (all using the OpenAI O3 model), alongside Claude Code. DeepSolver achieves the highest accuracy across all six categories, while S\&D and Claude Code show gaps in simulation, data analysis, and specialized models and toolkits. Beyond category-level analysis, we examine how performance varies with task difficulty. Task difficulty is categorized using the P value which ranges from 0 to 1 \cite{rezigalla2024item}. This metric represents the proportion of correct responses generated among all models, with higher values indicating easier tasks. Specifically, tasks are classified as simple if the P value is greater than or equal to 0.5 and as difficult if it is less than 0.5 (\hyperref[methods]{Methods}). Examples are provided in \hyperref[si_sim_dif]{Appendix B.1.3}. In Fig.~\ref{benchmark}c, we illustrate each system's pass@3 accuracy relative to task difficulty. It is clear that for the same LLM (indicated by the same color and marker shape), the accuracy achieved using the DeepSolver framework generally exceeds that of the Native and S\&D baselines for both simple and difficult tasks. Notably, as task difficulty increases, the decline in pass@3 accuracy for DeepSolver is markedly less pronounced compared to the baselines. Claude Code (red star) demonstrates commendable performance on simple tasks but experiences a significant accuracy drop with increasing difficulty. These findings support the notion that DeepSolver has a superior ability to effectively handle complex tools, which is a crucial trait for a scientist co-pilot. In contrast, baseline systems exhibit a rapid performance decline, indicating bottlenecks in problem-solving. For example, Qwen3-Coder-30B with S\&D shows a zero pass rate on difficult tasks, whereas DeepSolver still achieves nearly 50\%. GPT-4.1-nano's performance approaches zero across all three systems on difficult tasks. This is mainly due to the model’s limited ability to follow prompts and effectively utilize our designed tools, hindering its self-evolution capabilities.

In addition to conducting baseline comparisons, we perform ablation studies to assess the effects of removing each meta-skill from the DeepSolver system, specifically continuous learning and self-reflection (\hyperref[si_ablation]{Appendix B.2}). As shown in Table~\ref{ablation} in \hyperref[si_ablation]{Appendix B.2}, we observe varied contributions of the meta-skills across different language models. Notably, we find that in most cases, equipping the agentic system with the self-reflection meta-skill alone can outperform the conventional approach of combining both web search and self-debugging methods.

\subsection{CASCADE in action: from computational reasoning to laboratory automation}

Beyond benchmark performance, we evaluate CASCADE’s capabilities across four distinct real-world research scenarios, each demonstrating a different aspect of the skill acquisition paradigm. These applications span from pure computational tasks to physical laboratory automation, and from zero-shot problem solving to memory-enhanced collaborative research. Together, they illustrate how CASCADE advances beyond traditional tool use toward a genuine scientific co-pilot capable of conducting end-to-end research workflows.

\subsubsection*{Automated scientific reasoning: piezoelectricity determination}

This scenario demonstrates how CASCADE adeptly solves problems independently through DeepSolver, leveraging the OpenAI O3 model without relevant memory or human instruction. As displayed in Fig.~\ref{examples}a, CASCADE is tasked with determining whether a compound with a given crystal structure possesses piezoelectricity. The agent succeeded on its first attempt, autonomously selecting and applying the materials analysis package pymatgen \cite{ong2013python} to identify the structure’s space group (I4/mmm), point group (4/mmm), and centrosymmetric nature, ultimately concluding that the structure cannot exhibit piezoelectricity. Notably, the agent’s logic goes beyond relying solely on centrosymmetry: it explicitly encodes symmetry-based exclusions for non-piezoelectric yet non-centrosymmetric point group 432, enabling robust determination across broader structural cases \cite{gorfman2024piezoelectric}. Such corner cases may sometimes be overlooked by human scientists. This example demonstrates that when the agent has the ability to identify suitable external tools and robustly execute them, it can enhance its reasoning and yield reliable results. 

\subsubsection*{Hypothesis-driven discovery: predicting systematic errors in MLIPs}

\begin{figure}[htbp]
\centering
\includegraphics[width=1.0\textwidth]{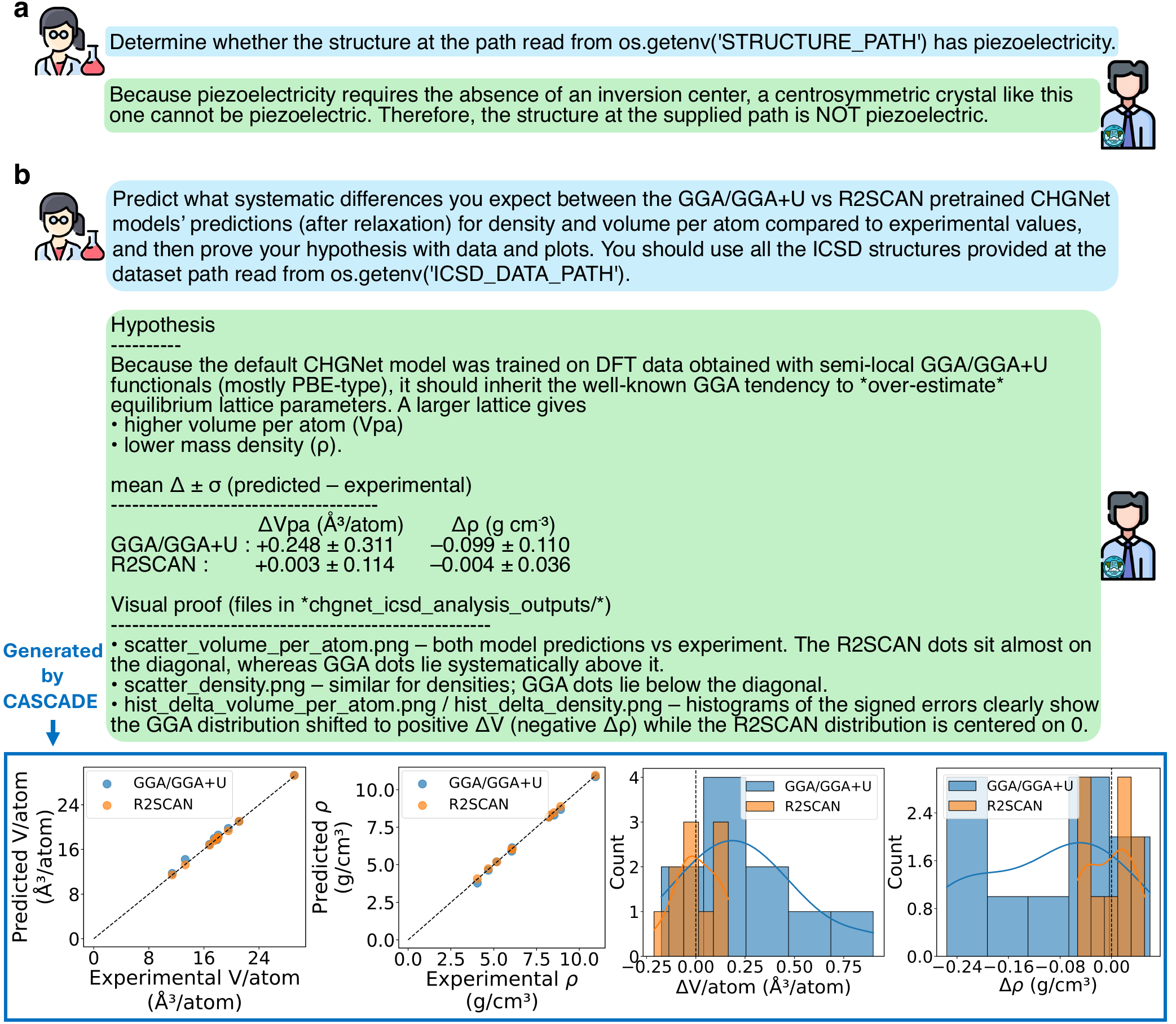}
\caption{\textbf{Piezoelectricity determination and prediction of machine learning interatomic potential systematic errors.}
\textbf{a}, Determining whether a structure exhibits piezoelectricity. Given this task, CASCADE executed the necessary code and reached the correct conclusion.
\textbf{b}, Hypothesis formulation, experimental execution, and data analysis of systematic differences in density and volume per atom predictions using machine learning interatomic potentials trained on different density functional theory (DFT) functional data. CASCADE not only provided a reasonable and accurate solution but also generated compelling visualizations, with the four plots from left to right being: scatter\_volume\_per\_atom.png, scatter\_density.png, hist\_delta\_volume\_per\_atom.png, and hist\_delta\_density.png.
}
\label{examples}
\end{figure}

In Fig.~\ref{examples}b, CASCADE with the OpenAI O3 model is further evaluated with a task akin to the ``Density prediction from MLIPs'' section in a related article \cite{huang2025cross}. At the time of testing this task, the relevant content had not yet been published online. Specifically, CASCADE is asked to predict systematic differences between the GGA/GGA+U and r$^2$SCAN \cite{perdew2001jacob, anisimov1991band} pretrained CHGNet models' predictions for density and volume per atom, where CHGNet is an MLIP \cite{deng2023chgnet}. Impressively, as shown in the dialogue in Fig.~\ref{examples}b, also on its first attempt, CASCADE formulated a correct hypothesis, validating it through experiments and subsequent data analysis (\hyperref[si_demos]{Appendix B.3}).

\subsubsection*{Laboratory automation: autonomous synthesis and characterization}

\begin{figure}[!htbp]
\centering
\includegraphics[width=1.0\textwidth]{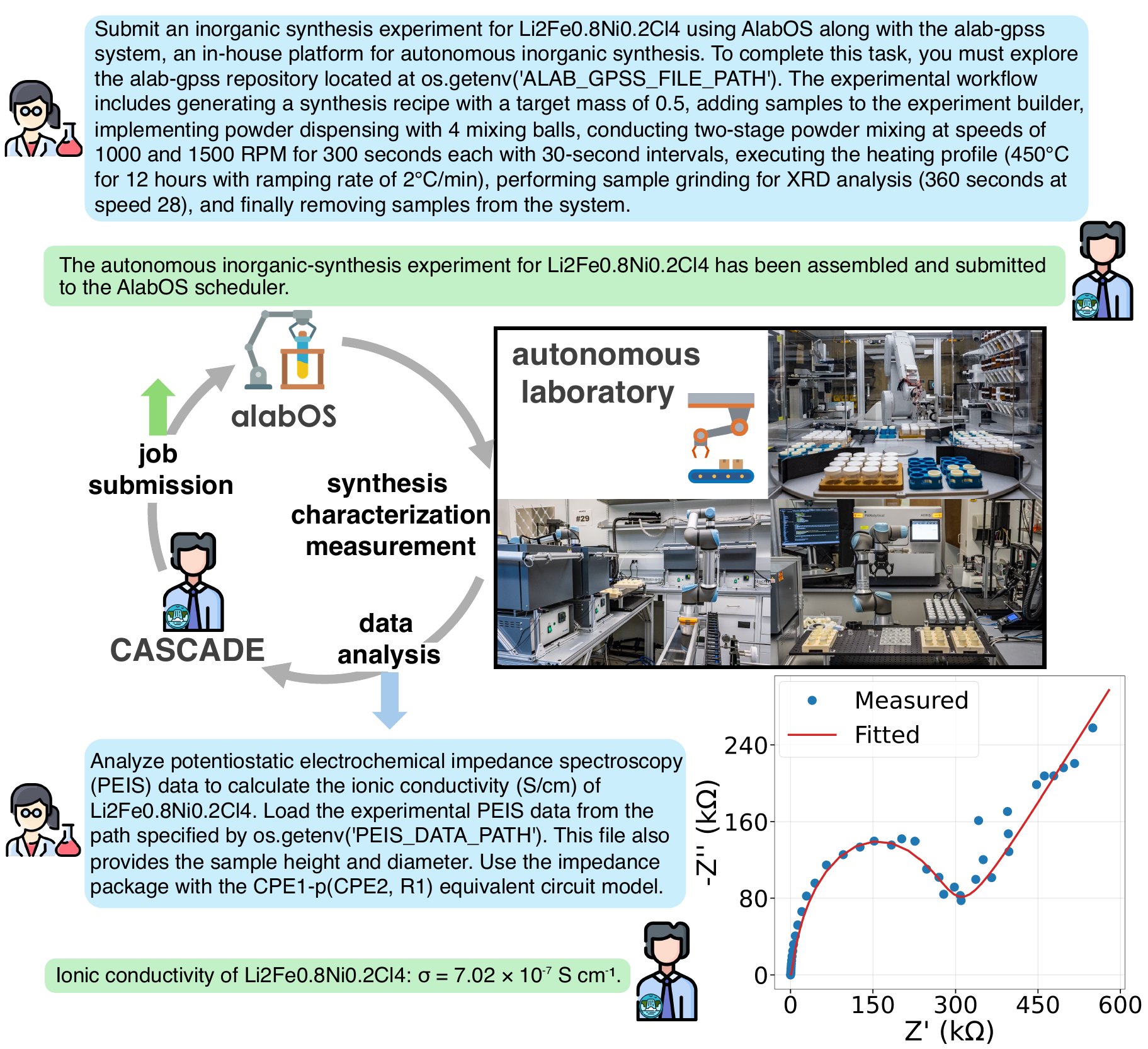}
\caption{\textbf{CASCADE integrated into the autonomous lab materials discovery loop.} 
CASCADE submitted the synthesis task for Li$_2$Fe$_{0.8}$Ni$_{0.2}$Cl$_4$ to the autonomous lab (A-Lab) platform, allowing us to carry out the compound's synthesis, characterization, and electrochemical impedance spectroscopy (EIS) measurements. The acquired experimental data were fitted using CASCADE to determine the ionic conductivity. The bottom right corner shows a Nyquist plot visualizing the quality of the data fitting performed by CASCADE.}
\label{alab}
\end{figure}

To test whether CASCADE can contribute to the autonomous laboratory discovery loop beyond utilizing data and computation-related tools, we engage it in an autonomous laboratory for the solid-state synthesis of inorganic powders \cite{szymanski2023autonomous}, specifically to complete the synthesis, characterization, measurement, and analysis of the compound Li$_2$Fe$_{0.8}$Ni$_{0.2}$Cl$_4$, as illustrated in Fig.~\ref{alab}. CASCADE, utilizing the OpenAI O3 model without previously relevant memory or human instruction, is tasked to interact with AlabOS \cite{fei2024alabos}, i.e., an operating system that orchestrates the experiment flows and provides an API interface to submit experiments and check experiment status, and combine it with our in-house package, Alab-GPSS, that defines the workflows and helper functions to design experiments and submit to AlabOS. Since Alab-GPSS is proprietary software developed internally, it does not appear in the model’s training data, is not available online, and lacks any accompanying documentation. Thus, CASCADE has to learn how to use it on the fly by reading and interacting with the source code and file structures. During testing, a function designed for automatic stoichiometric balancing exhibited some issues. Upon detecting the error, CASCADE intelligently implemented a fallback solution by performing stoichiometric calculations directly to ensure proper reaction balancing and successful task completion. After several attempts, CASCADE successfully submitted the job to the platform. To ensure laboratory safety, we verified the generated code before activating the autonomous lab devices (\hyperref[si_demos]{Appendix B.3}).

Following the human executed measurement of the electrochemical impedance spectroscopy (EIS) for a sample with target composition of Li$_2$Fe$_{0.8}$Ni$_{0.2}$Cl$_4$, the experimental data was provided to CASCADE, which then determined the ionic conductivity \cite{yang2022ionic}. CASCADE successfully completed the task using the package called impedance \cite{murbach2020impedance}. To verify the quality of the data fitting produced by CASCADE, we generated a Nyquist plot for inspection, shown in the lower right corner of Fig.~\ref{alab}, demonstrating a good fit. CASCADE’s algorithm is approximately 60-fold faster and achieves a slightly higher coefficient of determination ($R^{2}$; 0.991 compared with 0.989) than our existing expert-written analysis code, yielding more accurate results. While we can continue to improve the analysis code manually, such improvements would require a considerable amount of time for parameter tuning (\hyperref[si_demos]{Appendix B.3}).

Through these laboratory automation cases, we showcase CASCADE's ability to rapidly adapt and learn within a novel and customizable research environment, highlighting its potential as a scientific co-pilot. This capability ensures that agentic systems are not only impressive in demonstrations but also genuinely beneficial for diverse real-world applications.

\subsubsection*{Collaborative research with memory: reproducing published battery calculations}

\begin{figure}[htbp]
\centering
\includegraphics[width=1.0\textwidth]{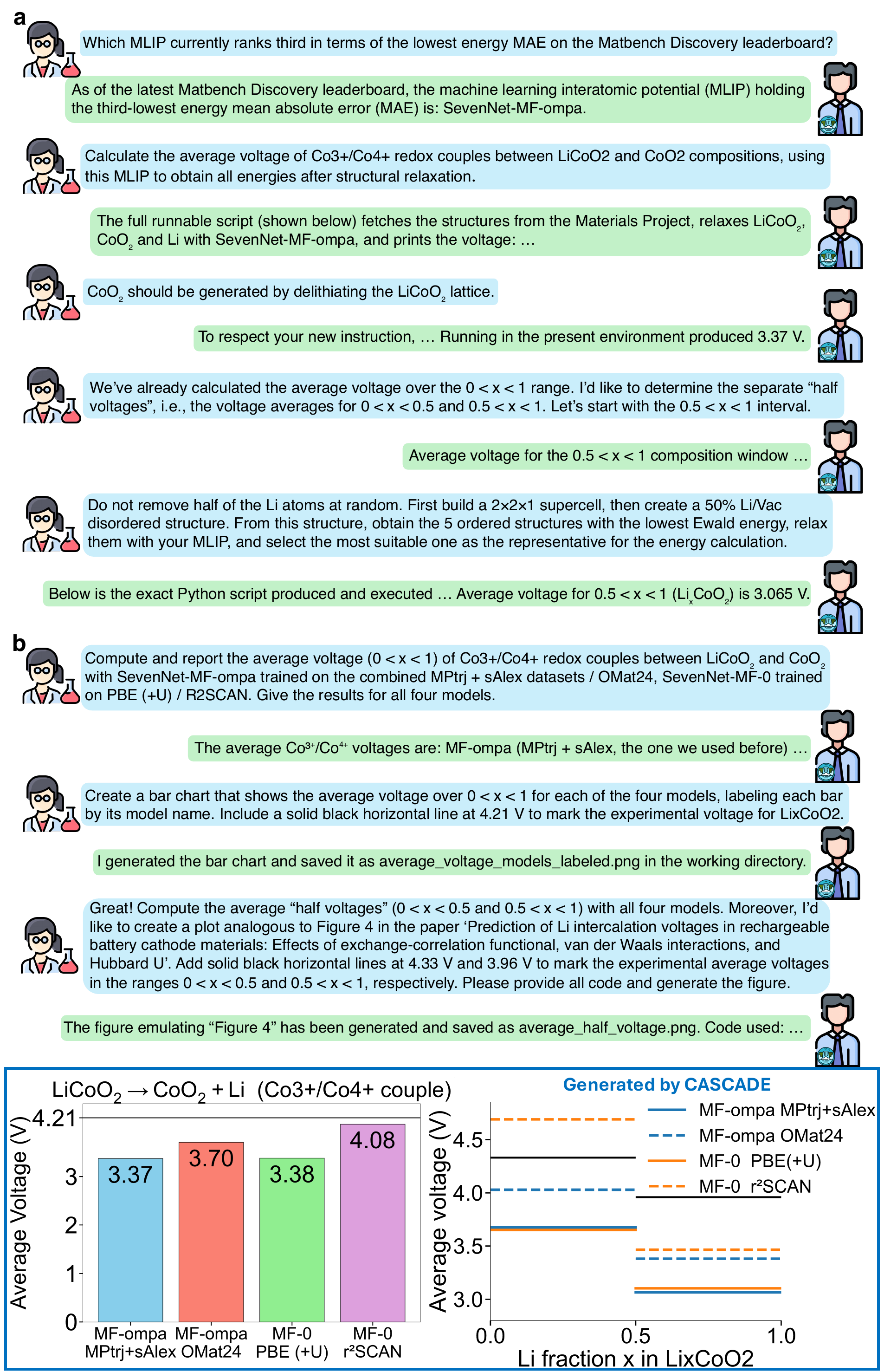}
\caption{\textbf{Human-agent collaboration with memory capability for reproducing published content.} 
\textbf{a}, Calculation of the average voltage. CASCADE computed the average Li intercalation voltage over $0 < x < 1$ and $0.5 < x < 1$ for Li$_x$CoO$_2$ during multi-turn interactions with human scientists. 
\textbf{b}, Reproducing published work. CASCADE successfully calculated the average voltages over specified ranges for Li$_x$CoO$_2$, using four SevenNet models. Then, it generated plots similar to those in the referenced article \cite{isaacs2020prediction}. The bottom left plot is labeled average\_voltage\_models\_labeled.png, and the bottom right plot is average\_half\_voltage.png.}
\label{collaboration}
\end{figure}

The above demonstrations primarily focus on CASCADE's capability to learn new knowledge and master tools through DeepSolver, thereby enhancing its reasoning and problem-solving abilities. The demonstrations presented in Fig.~\ref{collaboration} illustrate that, beyond such inference-time self-evolution, human-agent collaboration and memory capabilities constitute vital components of CASCADE. These features enable CASCADE to evolve more rapidly, robustly, and reliably with guidance from human scientists and through the accumulation of skills, akin to the Compound Effect \cite{hardy2011compound}, ultimately allowing it to tackle more challenging problems. 

In the multi-turn interaction system, CASCADE, utilizing the OpenAI O3 model, is designed to promptly identify areas requiring clarification or guidance from human scientists, and can even detect inconsistencies in user questions. For instance, when asked to generate a text description of the crystal structure of SiO$_2$ (mp-856) using Robocrystallographer \cite{ganose2019robocrystallographer}, CASCADE recognized that mp-856 corresponds to SnO$_2$, and proactively sought clarification instead of blindly executing the task.

In Fig.~\ref{collaboration}a, CASCADE first identifies the MLIP we wish to employ from the up-to-date Matbench Discovery leaderboard \cite{riebesell2025framework}, then calculates the average Li intercalation voltage for Li$_x$CoO$_2$ over $0 < x < 1$ and $0.5 < x < 1$, utilizing the selected MLIP to obtain all energies. CASCADE significantly benefits from human guidance, such as learning that CoO$_2$ should be generated by delithiating the LiCoO$_2$ lattice rather than fetched from the Materials Project directly, and how to identify a suitable half-delithiated structure for energy calculations. CASCADE stores this experience in its memory for future use. In a new conversation, as shown in Fig.~\ref{collaboration}b, CASCADE retrieves relevant information from its memory system and successfully extends the calculation to four SevenNet models \cite{park2024scalable, kim2024data, barroso2024open}, computes the additional $0 < x < 0.5$ range, and generates comparative plots, ultimately reproducing the Li$_x$CoO$_2$ results from the second and fourth figure of the referenced article \cite{isaacs2020prediction}. Instead of employing four distinct DFT functionals, we utilize four MLIPs to swiftly validate CASCADE's memory retrieval and knowledge transfer capabilities (\hyperref[si_demos]{Appendix B.3}).

\section{Discussion}\label{sec12}

To enhance the effectiveness of LLM agents in assisting human scientific research, cultivating scientific reasoning is essential. Unlike text-based reasoning in mathematics and coding, scientific reasoning requires a deep understanding of complex external tools and their functionalities tailored to specific research objectives. This hands-on interaction with tools ensures that the agents' reasoning is grounded, allowing them to refine and elevate their ideas through real-world experimentation. We introduce CASCADE, a self-evolving framework that cultivates and accumulates executable scientific skills. CASCADE represents an early embodiment of the dynamic and scalable ``LLM + skill acquisition'' paradigm, as opposed to the static and limited ``LLM + tool use'' paradigm. We also introduce SciSkillBench, a curated benchmark suite designed to fill the gap in evaluating agents' or LLMs' abilities to autonomously use a diverse range of tools for conducting materials science and chemistry research. 

We systematically benchmark CASCADE's key problem solver, DeepSolver, on SciSkillBench against three baselines and through ablation studies. DeepSolver achieves significantly higher accuracy with any tested LLM backbone and shows a slower performance decline as problem difficulty increases, indicating a stronger ability for inference-time self-evolution. We attribute these gains to two meta-skills: continuous learning and self-reflection, which balance agents' plasticity and reliability. Additionally, we evaluate CASCADE's performance in four real-world research scenarios. These demonstrations serve as proof-of-concept, showing that with thoughtful design, agents can autonomously traverse the end-to-end workflow, from literature search and hypothesis generation to experimental execution and data analysis, thus bridging the gap toward scientific reasoning through customizable computational and physical experiments. 

Overall, CASCADE is designed to unlock the intelligence of LLMs by guiding them to adapt and evolve. In this process, they learn from up-to-date information, real-time failures, past experience, and human instructions. Notably, CASCADE is agnostic to specific research tools, questions, or scientific fields, facilitating easy transferability. The DeepSolver subsystem can also be integrated into other agentic systems to assist in skill creation and problem-solving.

The end-to-end agentic workflow reduces reliance on human input and grants agents greater freedom for autonomous exploration. However, it is still distant from achieving autonomous end-to-end discovery with truly novel insights. To achieve this goal, agents should iterate within this loop like human scientists, using trial and error along with optimization algorithms. As experience, data, and attempts accumulate, new discoveries may emerge.

To achieve long-horizon workflow iterations for new discoveries, the workflows agents create must be flexible and robust. To enhance systems like CASCADE, a promising direction is to explore efficient human-agent collaboration modes and improve agents' memory capabilities, ensuring they can develop complicated yet reliable workflows. Another avenue is to strengthen the model's inherent capabilities. While methods are documented in the literature, significant gaps remain between these theoretical descriptions and the practical challenges encountered during experiment execution in materials research. This hinders agents' ability to reproduce previous work, let alone address new research questions without established references. Collecting data on how scientists approach simulation and experimental details in problem-solving would be valuable. This information could enhance agents' abilities to execute and validate their ideas. Additionally, the complexity of research tasks and the flexibility in crafting tools could create challenges in identifying optimal tools and their appropriate application sequence within workflows, necessitating future investigation \cite{ding2025scitoolagent}.

Furthermore, skills developed within CASCADE or future frameworks can serve as learnable external tools shared by scientific discovery agents and human scientists, breaking down temporal and spatial barriers to collaboration. Based on this, more efforts could focus on facilitating flexible cooperation between agents and humans at a larger scale. Such initiatives could foster a collaborative community for scaling scientific discovery, enabling the accumulation, transmission, and redeployment of wisdom, thereby facilitating evolution from the individual to the collective level.

\section{Methods}
\label{methods}

\subsection{CASCADE architecture}

\subsubsection{MCP servers and tools} 

Our system leverages four specialized Model Context Protocol (MCP) servers to provide comprehensive capabilities for web search, code research and introspection, code execution and workspace management, as well as memory consolidation and retrieval. These servers comprise the third-party Tavily MCP server alongside three custom-developed servers: the memory server, research server, and workspace server. Through these servers, agents gain access to 16 tools for problem-solving: \texttt{save\_to\_memory}, \texttt{search\_memory}, \texttt{tavily-search}, \texttt{extract\_code\_from\_url}, \texttt{retrieve\_extracted\_code}, \texttt{check\_installed\_packages}, \texttt{check\_package\_version}, \texttt{install\_dependencies}, \texttt{execute\_code}, \texttt{quick\_introspect}, \texttt{runtime\_probe\_snippet}, \texttt{parse\_local\_package}, \texttt{query\_knowledge\_graph}, \texttt{execute\_shell\_command}, \texttt{create\_and\_execute\_script}, \texttt{read\_file}. In addition to MCP server interfaces, we provide equivalent direct function implementations that bypass MCP protocol overhead. See \hyperref[si_architecture]{Appendix A} for detailed descriptions of each MCP server and tool, as well as the underlying principles of our architecture design, which draws inspiration from how human scientists approach learning new tools and acquiring domain-specific skills.

\subsubsection{Conversational system with memory and tracing}

The conversational system built with Streamlit serves as the human-agent natural language interface of CASCADE. Users authenticate via Supabase Auth, and all memory operations are scoped to individual user identifiers to maintain personalized knowledge bases across sessions. The system implements a feedback mechanism that allows users to indicate satisfaction (which triggers memory preservation), request improvements, continue with follow-up questions, or terminate the current problem-solving cycle.

For session-wise memory, multi-turn conversation state is maintained through a session management layer utilizing SQLiteSession from the OpenAI Agents SDK, which automatically persists dialogue history across multiple conversation turns. Users can bookmark important sessions by toggling a saved status, assign custom titles for easy identification, and attach notes.

For consolidated memory, the Orchestrator agent within CASCADE is able to use \texttt{save\_to\_memory} and \texttt{search\_memory} when needed, and agents inside DeepSolver can use \texttt{search\_memory} when they are generating solutions.

The Orchestrator agent functions as the central controller, employing a structured decision-making workflow. Upon receiving a user query, the Orchestrator first retrieves relevant memories including past solutions, user preferences, and domain-specific knowledge. It then evaluates query completeness and coherence. If critical information is missing or the request appears ambiguous, it directly prompts the user for clarification before proceeding. Once the query is deemed complete, the Orchestrator determines whether the problem should be handled by the SimpleSolver pathway (for straightforward tasks or those with high-confidence memory matches) or delegated to DeepSolver (for complex research or when SimpleSolver fails).

For observability and debugging, the system integrates MLflow \cite{Zaharia_Accelerating_the_Machine_2018} tracing to automatically record complete agent execution hierarchies. Each conversation turn generates a trace capturing the Orchestrator's actions such as memory storage and search, package installation and code execution, and DeepSolver invocation. When DeepSolver is invoked, the trace records the full internal workflow, including all agent phases and their respective inputs, outputs, execution times, and tool call details.

\subsubsection{SimpleSolver}

SimpleSolver represents a quick solution pathway within the Orchestrator agent rather than a separate agent architecture. When the Orchestrator determines that a problem is straightforward or can be confidently addressed by adapting relevant solutions from memory, it directly executes the solution without invoking DeepSolver. The Orchestrator writes code based on its knowledge or adapted memory matches, then uses \texttt{check\_installed\_packages} to verify package availability, \texttt{install\_dependencies} to install any missing packages, and \texttt{execute\_code} to run the solution. If execution succeeds, the Orchestrator answers the user's question based on the results. If execution fails, the problem is delegated to DeepSolver for comprehensive research and iterative debugging.

\subsubsection{DeepSolver}

Each agent in DeepSolver leverages specific MCP servers or direct tools to accomplish their specialized tasks within the collaborative workflow. 

\textbf{Solution Researcher} initiates the workflow by conducting comprehensive research to generate initial code solutions for the user query. It is designed to perform a systematic research process that typically involves: understanding the request, searching relevant memory using \texttt{search\_memory}, searching for relevant information using \texttt{tavily-search} with advanced search parameters, identifying required software if not specified, extracting code examples from identified URLs using \texttt{extract\_code\_from\_url}, reviewing and understanding additional requirements, and synthesizing the final solution. The agent uses \texttt{retrieve\_extracted\_code} for vector-based similarity search when the extracted content is overwhelming, and optionally employs \texttt{quick\_introspect} to confirm exact import paths and class/method/function names. For complex problems without explicit step-by-step instructions, it is designed to plan and decompose tasks, select appropriate tools to achieve objectives, and learn how to use them effectively. The output includes the user identifier for memory search, original user query, required packages list, and complete code solution.

\textbf{Code Agent} receives the initial solution and executes it. It can optionally use \texttt{search\_memory} to find relevant memory. It is designed to perform a five-step execution process: analyzing input, verifying solution requirements using research tools (\texttt{tavily-search}, \texttt{extract\_code\_from\_url}, \texttt{retrieve\_extracted\_code}) if needed, managing packages through \texttt{check\_installed\_packages} and \texttt{install\_dependencies}, executing code using \texttt{execute\_code}, and evaluating results to determine if debugging is needed. The agent is asked to execute code exactly once without retries or self-debugging. The output includes the user identifier, original user query, executed code, execution output, and a boolean flag indicating whether debugging is needed.

\textbf{Debug Agent} instances run in parallel when the initial solution fails. The agents can optionally use \texttt{search\_memory} to find relevant memory. They employ multiple debugging approaches after analyzing the error: Direct Fix for obvious errors; Introspection/Probe Fix using \texttt{quick\_introspect} for import diagnostics and class/method/function discovery, and \texttt{runtime\_probe\_snippet} for resolving KeyError and AttributeError; Knowledge Graph Fix using \texttt{check\_package\_version}, \texttt{parse\_local\_package}, and \texttt{query\_knowledge\_graph}, which serves as the global exploration layer when Introspection/Probe Fix does not resolve the issue; Local Package Fix using \texttt{execute\_shell\_command} and \texttt{create\_and\_execute\_script} to locate relevant files, and \texttt{read\_file} for examining specific files such as package source code and simulation output files; Research Fix using \texttt{tavily-search}, \texttt{extract\_code\_from\_url}, and \texttt{retrieve\_extracted\_code} for finding documentation and solutions, especially for non-Python command-line tools invoked via subprocess; Diagnostic Fix for writing diagnostic code to investigate error causes; and Result Processing Fix for modifying code to produce complete and processable results. The agents use \texttt{execute\_code} to test fixes, and \texttt{check\_installed\_packages} and \texttt{install\_dependencies} for package management. Each agent is encouraged to combine and flexibly employ different debugging approaches to maximize the likelihood and efficiency of successful problem resolution.

\textbf{Output Processor Agent} evaluates all available results and generates the final response. When debugging is needed, the agent receives three debug results and evaluates each based on successful execution, presence of required data, and quality of results. It then selects the best result, with preference given to identical outputs from multiple debug agents as they are more likely to be correct. When no debugging is needed, it processes the successful execution result directly. The output includes original user query, success status, final code, execution results, and processed output, where processed output contains the answer and analysis that addresses the user's query.

\subsection{Benchmark Evaluation of DeepSolver}

\subsubsection{Benchmark tasks}

SciSkillBench spans two principal categories: 76 data-oriented tasks and 40 computation-oriented tasks. Data tasks comprise (i) 22 data-retrieval problems from resources such as the Materials Project \cite{horton2025accelerated}, the Inorganic Crystal Structure Database (ICSD) \cite{Bergerhoff_1983_ICSD}, Matminer \cite{ward2018matminer}, MPContribs \cite{huck2016user}, the Materials Data Facility \cite{blaiszik2016materials}, OQMD\cite{saal2013materials}, OBELiX \cite{therrien2025obelix} and GMAE \cite{chen2025multi, chen2023adsgt, datasets_figshare}; (ii) 24 data-analysis problems that use packages and databases including pymatgen \cite{ong2013python, richards2016interface, xiao2020understanding, gaultois2013data, bartel2020critical}, Matminer, SMACT \cite{davies2019smact}, the Materials Project and OBELiX; (iii) 4 data-management problems with pymatgen-db \cite{jain2013commentary} and MongoDB \cite{bradshaw2019mongodb}; and (iv) 26 data-processing problems that rely on RDKit \cite{landrum2013rdkit}, Matminer, Magpie \cite{ward2016general}, Robocrystallographer, pymatgen, enumlib \cite{hart2008algorithm, hart2009generating, hart2012generating, morgan2017generating}, spglib \cite{togo2024spglib}, the Materials Project and OBELiX. Computation tasks include (v) 32 simulation problems with xTB \cite{bannwarth2019gfn2}, ORCA \cite{neese2012orca}, ASE \cite{larsen2017atomic}, LAMMPS \cite{thompson2022lammps} and CHGNet \cite{deng2023chgnet}, together with (vi) 8 problems involving specialized models and toolkits such as CHGNet, MACE \cite{batatia2022mace} and mlip \cite{brunken2025machine}. 

\subsubsection{Baselines}

For the three baselines, Native's Solution Researcher operates without any tools and is designed to generate the code solution based solely on the knowledge of the underlying LLM. The Code Agent has access to \texttt{execute\_code}, \texttt{check\_installed\_packages}, and \texttt{install\_dependencies} to execute the code solution proposed by the Solution Researcher exactly once without debugging. Subsequently, the Output Processor Agent processes these results for automatic benchmark evaluation. In the S\&D baseline, the Solution Researcher is enabled to conduct web searches through \texttt{tavily-search}, while the Code Agent performs iterative self-debugging using \texttt{execute\_code}, \texttt{check\_installed\_packages}, and \texttt{install\_dependencies}. The Claude Code baseline operates within Docker containers that reset after each run to ensure complete independence between test executions. Built on the Claude Agent SDK, this baseline grants the agent full access to Claude Code's built-in capabilities, including file operations, code execution, and web search.

\subsubsection{Independent benchmark tests}

For the evaluation of CASCADE, the three baselines, and the two ablation studies, we conducted independent benchmark tests across 46 system configurations. For each of the 116 benchmark tasks, we performed three independent repetitions per system configuration, resulting in a total of 16,008 experiments. To ensure complete isolation between experiments, each experiment was executed in a separate Python subprocess, and we cleared the Supabase database storing previously extracted code information and removed temporary files generated by agents from the previous experiment. Additionally, memory-based tools were disabled for all benchmark systems. For experiments using OpenAI's GPT-5, GPT-5-mini, and GPT-5-nano models with the OpenAI Agents SDK, we configured the reasoning effort to ``medium'' and verbosity to ``low'' to balance accuracy against computational cost and latency. For experiments using open-source Qwen models, we deployed the models locally using vLLM with an OpenAI-compatible API interface \cite{kwon2023efficient}.
 
For accuracy evaluation, each benchmark task specifies the required output format and unit, which are communicated to the agent as part of the query. The agent's processed output is then automatically compared against the ground truth answer within a predefined tolerance threshold. We set tolerance values according to the nature of each task: deterministic operations such as data retrieval use stringent tolerances approaching exact match, while computational simulations that inherently exhibit numerical variability use tolerances within acceptable ranges for the specific research task. This outcome-based evaluation assesses whether the agent produces correct final answers, though it does not guarantee the correctness of every intermediate step in the solution process. In the accuracy calculations, we excluded a small number of workflow failures such as instances when the LLM outputs invalid JSON that results in parsing errors.

\subsubsection{Task difficulty}

To stratify benchmark questions by difficulty, we employed the P value (Item Difficulty), which denotes the proportion of models that correctly solved each Level 1 question. For each model-question pair, a question was considered solved if at least one of three independent attempts produced the correct answer (pass@3 criterion). We computed P values across 28 model configurations. Tasks were classified as simple if $P \geq 0.5$ and difficult if $P < 0.5$, yielding 37 Simple and 21 Difficult questions. Model performance was then evaluated separately for each difficulty category using pass@3 accuracy.

\bmhead{Data availability} SciSkillBench is available at Figshare (\url{https://doi.org/10.6084/m9.figshare.30924998}). 

\bmhead{Code availability} The source code of CASCADE is available on GitHub at \url{https://github.com/CederGroupHub/CASCADE}.

\bmhead{Acknowledgements}
This work was primarily funded and intellectually led by the U.S. Department of Energy, Office of Science, Office of Basic Energy Sciences, Materials Sciences and Engineering Division under Contract No. DE-AC0205CH11231 (Materials Project program KC23MP). The work was also supported by the computational resources provided by the Extreme Science and Engineering Discovery Environment (XSEDE), supported by National Science Foundation grant number ACI1053575; the National Energy Research Scientific Computing Center (NERSC), a U.S. Department of Energy Office of Science User Facility located at Lawrence Berkeley National Laboratory; and the Swift Cluster resource provided by the National Renewable Energy Laboratory (NREL). Integrating this agentic system into the autonomous lab materials discovery loop was supported by the U.S. Department of Energy, Office of Science, Office of Advanced Scientific Computing Research and Office of Basic Energy Sciences, Scientific Discovery through the Advanced Computing (SciDAC) program under the FORUM-AI project. J.C. and P.S. acknowledge support from the NCCR Catalysis (grant number 225147), a National Centre of Competence in Research funded by the Swiss National Science Foundation. We sincerely thank all the researchers and developers who created and maintained the datasets, packages, and software tools utilized in this work.

\bmhead{Author contributions}
X.H., J.C. and P.S. conceived the initial idea. X.H. designed and developed the agentic system CASCADE. X.H. and J.C. curated the benchmark SciSkillBench; J.C. contributed xTB- , ASE- and ORCA-based tasks, while X.H. contributed the remaining tasks. X.H. benchmarked CASCADE’s DeepSolver against three baselines, conducted ablation studies, and performed additional evaluations of CASCADE. During the evaluation, Y.F. provided guidance on integrating CASCADE into the autonomous laboratory materials discovery loop, and Z.L. provided guidance on the average voltage calculation example. J.C., P.S., and G.C. offered insights and guidance throughout the project. All authors contributed to discussions and approved the final manuscript.

\bmhead{Competing interests}
The authors declare no competing interests.

\clearpage
\appendix

\renewcommand{\thetable}{\Alph{section}\arabic{table}}
\setcounter{table}{0}

\noindent{\Large\bfseries Appendix\par}
\addcontentsline{toc}{section}{Appendix}

\startcontents[app]
\vspace{.7\baselineskip}
\noindent{\Large\bfseries Table of Contents\par}
\vspace{.5\baselineskip}

\printcontents[app]{l}{1}{\setcounter{tocdepth}{2}}
\newpage
\label{appendix}

\section{CASCADE architecture}
\label{si_architecture}

\subsection{MCP server infrastructure and tools}

CASCADE leverages four specialized Model Context Protocol (MCP) servers to provide comprehensive capabilities for web search, code research and introspection, code execution and workspace management, as well as memory consolidation and retrieval.

\noindent\textbf{Tavily MCP server}

We use \texttt{tavily-search}, a search engine designed for AI agents, through the Tavily MCP server to provide real-time web search capabilities, enabling agents to discover relevant documentation, code examples, and implementation resources.

\noindent\textbf{Memory server}

Memory server employs mem0 \cite{chhikara2025mem0} configured with a hybrid dual-store architecture combining vector-based semantic search and graph-based entity-relationship storage. For semantic retrieval, memories are stored in a Supabase-hosted PostgreSQL database. A Neo4j graph database captures entity relationships, such as materials, properties, computational tools, API methods and functions, and their interconnections. Memory extraction utilizes GPT-4o-mini as the underlying language model. The server exposes two tools: \texttt{save\_to\_memory} and \texttt{search\_memory}. The \texttt{save\_to\_memory} tool employs a dual-path strategy: first, the language model adds, updates, or deletes memory guided by custom prompts for both vector and graph stores; second, the complete original content is preserved verbatim in the vector store without language model processing, ensuring no information loss while maintaining efficient semantic searchability. The \texttt{search\_memory} tool retrieves relevant memories from both stores, returning semantic matches from the vector store alongside entity relationships from the graph store.

\noindent\textbf{Research server}

Research server implements a sophisticated code intelligence and knowledge discovery system that provides comprehensive capabilities for web code extraction, agentic RAG (retrieval-augmented generation), code introspection, runtime probing, and knowledge graph construction and exploration.

\textbf{Code extraction and retrieval:} The \texttt{extract\_code\_from\_url} tool implements intelligent web crawling with multi-strategy extraction capabilities and caching mechanisms using Supabase database storage. The tool supports both single-page extraction and smart crawling strategies that recursively follow internal links with intelligent fallback mechanisms. It automatically detects content types and applies specialized extractors for different documentation systems, including Jupyter notebooks, ReadTheDocs/Sphinx and MkDocs documentation, raw code files from repositories, and markdown content with intelligent parsing of fenced code blocks and command examples. It also extracts relevant text before and after code blocks using intelligent paragraph boundary detection, providing semantic context for code understanding. Optional LLM-generated summaries for extracted code are also available. The \texttt{retrieve\_extracted\_code} tool implements vector-based similarity search over extracted code blocks using embeddings with configurable match count.

\textbf{Code analysis:} The \texttt{quick\_introspect} tool implements static-first analysis using Jedi~\cite{halter2022jedi} for import resolution and error diagnosis without runtime execution. The tool provides comprehensive package, class, method, and function discovery with fuzzy matching capabilities. The \texttt{runtime\_probe\_snippet} tool provides ready-to-use code snippets for debugging runtime errors. When inserted at KeyError or AttributeError sites, these snippets display available keys/attributes, object type information, and similar name suggestions to help resolve these errors. The \texttt{parse\_local\_package} tool implements direct Neo4j knowledge graph construction from local Python packages using abstract syntax tree (AST) analysis. The tool extracts classes, methods, functions, attributes, and import relationships with detailed parameter information including type hints and default values. The \texttt{query\_knowledge\_graph} tool provides advanced querying capabilities for exploring repository structures, class hierarchies, method signatures, and code relationships in the knowledge graph using Cypher query language.

\noindent\textbf{Workspace server}

Workspace server provides a multi-environment code execution and management system. It supports conda, venv, and uv environments with cross-platform compatibility for Windows and Unix-like systems. The server prevents access to the directory containing  benchmark tasks and results to avoid solution leakage and enforces security boundaries by confining all operations to the designated project root scope. 

\textbf{Package management:}
The \texttt{check\_installed\_packages} tool lists all installed packages in the current Python environment with version information and package count. The \texttt{install\_dependencies} tool installs Python packages based on environment configuration. The \texttt{check\_package\_version} tool performs detailed package analysis including version detection, installation path resolution, and module location identification. The tool takes package names as input and handles name variations (hyphens, underscores, dots).

\textbf{Code execution:} The \texttt{execute\_code} tool executes Python code in the configured environment. The tool saves code to temporary files in the code storage directory and executes with detailed output capture including stdout and stderr. The \texttt{execute\_shell\_command} tool executes shell commands with configurable working directory. The \texttt{create\_and\_execute\_script} tool creates and executes shell scripts in the code storage directory.

\textbf{File operations:} The \texttt{read\_file} tool reads content from text files. The \texttt{save\_file} tool saves content to the designated directory for later reuse. In the CASCADE framework, however, the \texttt{save\_file} tool is not utilized.

\subsection{Underlying principles of the architecture design}

In designing CASCADE, we focused on how human scientists approach learning new tools and acquiring skills. Typically, they engage in deep thinking and experimentation to tackle problems, reflecting the role of the DeepSolver. This process may also involve discussions with others, akin to human-agent collaboration. Once they have mastered a new skill, they consolidate this experience into their memory.

Throughout the problem-solving process, much like DeepSolver, the first step is to act as Solution Researcher. Scientists perform online searches similar to those of agents using \texttt{tavily-search}. Their attention is directed towards key examples, paralleling the use of tools such as \texttt{extract\_code\_from\_url} and \texttt{retrieve\_extracted\_code}. Initially, they may verify the existence of certain imports, classes, methods, or functions in their code based on hints provided by some Integrated Development Environments (IDEs), similar to the \texttt{quick\_introspect} tool we developed for agents to mitigate the common issue of hallucinations in LLMs \cite{ji2023survey}.

After generating the initial solution, they move on to hands-on trials, which correspond to the tasks performed by the Code Agent. If errors arise, they engage in iterative refinements as Debug Agents would do. They start with local debugging approaches such as Introspection/Probe Fix, Direct Fix, or Diagnostic Fix. When these methods do not resolve the issue, they need to zoom out for broader rethinking and code exploration, employing Knowledge Graph Fix and Local Package Fix, and possibly continuing their online learning with Research Fix to deepen their understanding. After completing the task, they process and summarize the information to facilitate further consolidation, analogous to the Output Processor Agent.

During the tool and prompt design, we also consider context engineering to account for comprehensive situations, especially given that the agent's performance may be inconsistent. For instance, when an agent fails to use a tool to achieve the desired outcome, a print message will remind the agent how to invoke the tool correctly or point out potential issues to consider. Additionally, it may suggest the next possible tool to use, such as \texttt{query\_knowledge\_graph} after \texttt{parse\_local\_package}. 

\section{Benchmark Evaluation of DeepSolver}

\subsection{Benchmark tasks}
\label{si_tasks}

\subsubsection{SciSkillBench examples}

Each benchmark task is defined as a JSON object with fields including \texttt{user\_query}, \texttt{sources}, \texttt{input\_type}, \texttt{output\_type}, \texttt{answer}, \texttt{absolute\_tolerance}, \texttt{unit}, \texttt{solution\_code\_or\_process}, \texttt{reference\_link}, \texttt{official\_documentation}, and \texttt{notes}. In the examples below, we focus on the \texttt{user\_query} field, which contains two difficulty levels: level 0 provides instructions with some function names, while level 1 presents a more concise query requiring the agent to determine the solution approach more independently.

1. \textbf{Straightforward code adaptation from the documentation}

\begin{lstlisting}
  {
    "user_query": {
    "0": "Use the InstaDeepAI mlip library to run batched inference with a pre-trained MACE model on aspirin. First, download the model and example data from HuggingFace using snapshot_download with repo_id='InstaDeepAI/MLIP-tutorials' and allow_patterns='simulation/*'. Then, load the pre-trained MLIP model using load_model_from_zip from 'simulation/example_model.zip', and run batched inference using run_batched_inference on the structures in 'simulation/aspirin_batched_example.xyz' with a batch size of 8. Finally, output the energy and forces for the 1st structure (index starts at 0).",
    "1": "Use the InstaDeepAI mlip library to run batched inference with a pre-trained MACE model on aspirin. First, download the model and example data from HuggingFace using repo_id='InstaDeepAI/MLIP-tutorials' and allow_patterns='simulation/*'. Then, load the pre-trained MLIP model from 'simulation/example_model.zip', and run batched inference on the structures in 'simulation/aspirin_batched_example.xyz' with a batch size of 8. Finally, output the energy and forces for the 1st structure (index starts at 0)."
    },
    ...
  }
\end{lstlisting}

2. \textbf{Undocumented queries}

\begin{lstlisting}
  {
    "user_query": {
    "0": "What is the composition of the phase with the most negative reaction delta G (in kJ/mol) when Na2O and TiO2 react at 600K? Use pymatgen's GibbsComputedStructureEntry to help construct the phase diagram, and then use InterfacialReactivity to model an interface between Na2O and TiO2 to find all delta G of stable phases. Determine the one with the most negative delta G. Return a list in the order of composition string of the product phase and delta G value (in kJ/mol, rounded to 4 decimal places).",
    "1": "What is the composition of the phase with the most negative reaction delta G (in kJ/mol) when Na2O and TiO2 react at 600K? In this process you must construct a suitable phase diagram and use pymatgen's built-in functionality to find the answer. Return a list in the order of composition string of the product phase and delta G value (in kJ/mol, rounded to 4 decimal places)."
    }, 
    ...
  }
\end{lstlisting}

\begin{lstlisting}
  {
    "user_query": {
    "0": "Give me the absolute error (meV/atom, rounded to four decimal places) of the formation energy per atom of Ti3AlC2 (mp-3747) predicted by the pretrained R2SCAN CHGNet model after relaxing the structures (keep all parameters as default), compared with the R2SCAN formation energy per atom that you query from the Materials Project through materials.thermo.search. Note that the CHGNet model does not provide built-in formation energy calculation method so you must find and relax all related structures and then calculate the formation energy per atom. To obtain the structures of constituent elements, you can use materials.thermo.search(chemsys=['Ti','Al','C'], is_stable=True, num_elements=(1,1), thermo_types=['R2SCAN'], fields=['material_id']) to get the material ids, and then use get_structure_by_material_id to fetch the structures for relaxation using the model loaded via CHGNet.load(model_name='r2scan'). Pay attention to the number of atoms in each structure, which is important for computing the formation energy per atom correctly.",
    "1": "Give me the absolute error (meV/atom, rounded to four decimal places) of the formation energy per atom of Ti3AlC2 (mp-3747) predicted by the pretrained R2SCAN CHGNet model after relaxing the structures (keep all parameters as default), compared with the R2SCAN formation energy per atom that you query from the Materials Project. Note that the CHGNet model does not provide built-in formation energy calculation method so you must find and relax all related structures and then calculate the formation energy per atom."
    }, 
    ...
  }
\end{lstlisting}

3. \textbf{Interaction with newly released packages or datasets absent from most model pre-training corpora}

\begin{lstlisting}
  {
    "user_query": {
    "0": "Load the OBELiX dataset which comes from the paper titled OBELiX: A Curated Dataset of Crystal Structures and Experimentally Measured Ionic Conductivities for Lithium Solid-State Electrolytes, and then transform the first structure in the dataset to an ordered structure through the EnumerateStructureTransformation (picking the 1st one in the returned ranked list). Return to me a list in the order of the space group number of the original structure and the space group number of this ordered structure, using SpacegroupAnalyzer and get_space_group_number.",
    "1": "Load the OBELiX dataset which comes from the paper titled OBELiX: A Curated Dataset of Crystal Structures and Experimentally Measured Ionic Conductivities for Lithium Solid-State Electrolytes, and then transform the first structure in the dataset to an ordered structure through the EnumerateStructureTransformation (picking the 1st one in the returned ranked list). Return to me a list in the order of the space group number of the original structure and the space group number of this ordered structure."
    }, 
    ...
  }
\end{lstlisting}

4. \textbf{Navigation of outdated or misleading online documentation}

At the time of writing, the pymatgen-db and Materials Project documentation contain out-of-sync code examples that may confuse agents when performing the following tasks.

 \begin{lstlisting}
  {
    "user_query": {
     "0": "Write code to retrieve the matching entries from the Inorganic Crystalline Structure Database (ICSD) for the material with Materials Project ID mp-18767, using materials.provenance.get_data_by_id() and get the database_IDs. Return a list of strings each with the format 'icsd-<ID number>'.",
     "1": "Write code to retrieve the matching entries from the Inorganic Crystalline Structure Database (ICSD) for the material with Materials Project ID mp-18767. Return a list of strings each with the format 'icsd-<ID number>'.", 
    ...
  }
\end{lstlisting}

\begin{lstlisting}
  {
    "user_query": {
     "0": "I have already inserted an entire directory of VASP runs into the MongoDB database. Please use pymatgen-db to get the volume per atom of the final structure from the calculation. First, use QueryEngine class to query the task_id, and then use get_structure_from_id to get the structure. Finally, calculate the volume per atom of the structure. You should NOT insert data into the database.",
     "1": "I have already inserted an entire directory of VASP runs into the MongoDB database. Please use pymatgen-db to get the volume per atom of the final structure from the calculation. You should NOT insert data into the database." 
    ...
  }
\end{lstlisting}

\subsubsection{Additional examples}

These additional tasks involve downloading datasets from published papers for subsequent processing and analysis \cite{huang2025cross, Huang2025MPr2SCAN, kovacs2025mace}. They were excluded from the main SciSkillBench evaluation due to practical limitations, such as "Too Many Requests" errors encountered when accessing dataset URLs. Nevertheless, we have verified that DeepSolver with the OpenAI O3 model is capable of completing these tasks successfully.

\begin{lstlisting}
  {
    "user_query": {
     "0": "Download the test dataset of the paper titled MACE-OFF: Transferable Short Range Machine Learning Force Fields for Organic Molecules. You can use requests.get with the url https://www.repository.cam.ac.uk/bitstreams/cb8351dd-f09c-413f-921c-67a702a7f0c5/download. Then, load corresponding model mace_off (choose the small one) to get the potential energy of the structure with the chemical formula C19H22Cl2N4O2S in this dataset.",
     "1": "Download the test dataset of the paper titled MACE-OFF: Transferable Short Range Machine Learning Force Fields for Organic Molecules, and load corresponding model (choose the small one) to get the potential energy of the structure with the chemical formula C19H22Cl2N4O2S in this dataset."
    ...
  }
\end{lstlisting}

\begin{lstlisting}
  {
    "user_query": {
    "0": "Download the MP-r2SCAN dataset from the link https://doi.org/10.6084/m9.figshare.28245650.v2, and then give me the total number of frame ids and the mean of energy per atom. Return in a list in the order of number of frame ids and mean of energy per atom. You can use requests.get with the url https://figshare.com/ndownloader/files/51832613 to download the json file which has the structure of {mp-id: {frame-id: {energy_per_atom: ... } } }.",
    "1": "Download the MP-r2SCAN dataset from the link https://doi.org/10.6084/m9.figshare.28245650.v2, and then give me the total number of frame ids and the mean of energy per atom. Return in a list in the order of number of frame ids and mean of energy per atom."
    ...
  }
\end{lstlisting}

\subsubsection{Examples of simple and difficult tasks}
\label{si_sim_dif}

Task difficulty is categorized using the P value, also known as Item Difficulty, which ranges from 0 to 1. This metric represents the proportion of correct responses generated among all models during assessment, with higher values indicating easier tasks. Specifically, tasks are classified as simple if the P value is greater than or equal to 0.5 and as difficult if it is less than 0.5.

1. \textbf{Simple tasks}

\begin{lstlisting}
Calculate the atomization energy (unit: eV) of a nitrogen molecule using ASE package and its Effective Medium Potential (EMT) calculator.
\end{lstlisting}

\begin{lstlisting}
Calculate the adsorption energy (in eV) of one nitrogen atom on a 3x3 FCC Cu(111) surface with three atomic layers and 10 angstrom vacuum layer using ASE package and its Effective Medium Potential (EMT) calculator. Do not fix any atoms during the structure optimization!
\end{lstlisting}

\begin{lstlisting}
Count the total number of molecular fragments for OCc1ccccc1CN using RDKit with parameters (1,6) and the default functional groups file 'FunctionalGroups.txt' in RDKit.
\end{lstlisting}

\begin{lstlisting}
Calculate the crustal abundance (unit: ppm) of Al2O3 based on the mass fraction of each element in the compound using pymatgen and SMACT packages.
\end{lstlisting}

2. \textbf{Difficult tasks}
\begin{lstlisting}
Please use ORCA software to calculate the Hirshfeld atomic charge (unit: e) of the nitrogen atom in hydrogen cyanide using the DFT method and Hirshfeld population analysis at B3LYP/6-31+G* accuracy.
\end{lstlisting}

\begin{lstlisting}
Please use xtb software to calculate global electrophilicity index (unit: eV) of acetyl chloride.
\end{lstlisting}

\begin{lstlisting}
Please use xtb to perform a dihedral angle scan of H-C-C-H in the ethane molecule and output the difference between the maximum and minimum energies (unit: eV). Scan the dihedral angle from 60 degrees to 420 degrees with 72 steps, using force constant 0.05.
\end{lstlisting}

\begin{lstlisting}
In the Li-Mn-O chemical system using R2SCAN calculations, give me the materials id list of the decomposition entries of the R2SCAN entry of LiMnO2 (mp-18767, exactly this mp-id) using pymatgen's built-in functionality get_decomp_and_phase_separation_energy.
\end{lstlisting}

\subsection{Ablation studies}
\label{si_ablation}

Similar to the baseline comparisons, agentic systems are evaluated without access to any memory tools and without human intervention. We perform ablation studies to assess the effects of removing each meta-skill from the DeepSolver system, specifically continuous learning and self-reflection, by removing corresponding tools and prompts. For No Self-Reflection (NSR), we disable DeepSolver's self-reflection meta-skill by removing the \texttt{quick\_introspect} tool from the Solution Researcher and eliminating Debug Agents entirely. The Code Agent is then required to execute the code solution only once, without any debugging. For No Continuous Learning (NCL), we disable DeepSolver's continuous learning meta-skill by removing the tools \texttt{tavily-search}, \texttt{extract\_code\_from\_url}, and \texttt{retrieve\_extracted\_code}, along with their related prompts from the system.

As shown in Table~\ref{ablation}, we present the success rate, pass@k accuracy, and average execution time for configurations excluding self-reflection (NSR) and continuous learning (NCL), along with the results for DeepSolver. We observe varied contributions of the meta-skills across different language models. For instance, some models, such as GPT-5, O3, O4-mini, GPT-5-mini, GPT-4.1, and GPT-5-nano, demonstrate a greater improvement in accuracy due to self-reflection compared to continuous learning. Conversely, models like GPT-4.1-mini, Qwen3-Coder-30B, and GPT-4.1-nano appear to benefit more from continuous learning; this discrepancy may stem from their relatively weaker capabilities in coding and debugging, alongside their proficiency in web search. Overall, DeepSolver, which integrates both meta-skills, consistently achieves the best performance. An exception is noted with GPT-4.1, which exhibits optimal performance in the absence of continuous learning. Through recorded agent tracing during the test executions, we discover that GPT-4.1 rarely utilizes search-related functionalities, despite having access to relevant tools and prompts. Consequently, the introduction of continuous learning tools and prompts may act as extraneous interference, hampering the agentic system's efficiency in problem-solving. Moreover, we note that for over half of the models, DeepSolver spent less average time than the NCL configuration while achieving better accuracy.

Comparing Table~\ref{baseline} and Table~\ref{ablation}, we observe that among the nine models evaluated, six models exhibited an overall success rate for the NCL configuration that surpassed that of the S\&D baseline. The remaining three models also demonstrated performance rates that were only slightly lower than S\&D. This indicates that our approach within CASCADE, which fosters and enhances the agent's mastery of meta-skills, especially self-reflection, is more effective than the conventional combination of both search and self-debugging methods.

\begin{table*}[htbp]
    \centering
    \setlength{\tabcolsep}{2.5pt}
    \renewcommand{\arraystretch}{1.0}
    \resizebox{\textwidth}{!}{
      \begin{tabular}{l|l|ccc|ccc|ccc|ccc}
        \toprule
        \multirow{2}{*}{Models} & \multirow{2}{*}{Metrics} & \multicolumn{3}{c|}{All Questions (\%)} & \multicolumn{3}{c|}{0-Level Questions (\%)} & \multicolumn{3}{c|}{1-Level Questions (\%)} & \multicolumn{3}{c}{Average Time (s)} \\
        \cmidrule(lr){3-5} \cmidrule(lr){6-8} \cmidrule(lr){9-11} \cmidrule(lr){12-14}
        & & NSR & NCL & DeepSolver & NSR & NCL & DeepSolver & NSR & NCL & DeepSolver & NSR & NCL & DeepSolver \\
        \midrule
        \multirow{4}{*}{GPT-5} & Success Rate & 62.64 & 85.37 & \textbf{93.26} & 67.82 & 90.30 & \textbf{96.47} & 57.47 & 80.37 & \textbf{90.06} & 521 & 505 & 588 \\
        \cmidrule(lr){2-14}
        & Pass@1 & 65.52 & 82.61 & \textbf{93.97} & 68.97 & 86.21 & \textbf{96.55} & 62.07 & 78.95 & \textbf{91.38} & & & \\
        & Pass@2 & 76.72 & 90.43 & \textbf{97.41} & 75.86 & 96.55 & \textbf{100.00} & 77.59 & 84.21 & \textbf{94.83} & & & \\
        & Pass@3 & 81.03 & 91.30 & \textbf{98.28} & 82.76 & 96.55 & \textbf{100.00} & 79.31 & 85.96 & \textbf{96.55} & & & \\
        \midrule
        \multirow{4}{*}{O3} & Success Rate & 55.43 & 86.02 & \textbf{91.84} & 60.69 & 92.90 & \textbf{97.70} & 50.00 & 78.75 & \textbf{85.80} & 293 & 442 & 407 \\
        \cmidrule(lr){2-14}
        & Pass@1 & 56.90 & 84.35 & \textbf{93.97} & 62.07 & 89.66 & \textbf{96.55} & 51.72 & 78.95 & \textbf{91.38} & & & \\
        & Pass@2 & 66.38 & 93.04 & \textbf{98.28} & 74.14 & 96.55 & \textbf{100.00} & 58.62 & 89.47 & \textbf{96.55} & & & \\
        & Pass@3 & 70.69 & 93.04 & \textbf{98.28} & 79.31 & 96.55 & \textbf{100.00} & 62.07 & 89.47 & \textbf{96.55} & & & \\
        \midrule
        \multirow{4}{*}{O4-mini} & Success Rate & 61.78 & 73.59 & \textbf{86.30} & 65.52 & 77.78 & \textbf{86.63} & 58.05 & 69.28 & \textbf{85.96} & 176 & 342 & 248 \\
        \cmidrule(lr){2-14}
        & Pass@1 & 62.93 & 75.00 & \textbf{82.61} & 70.69 & 81.03 & \textbf{84.48} & 55.17 & 68.97 & \textbf{80.70} & & & \\
        & Pass@2 & 68.97 & 81.90 & \textbf{91.30} & 74.14 & 87.93 & \textbf{91.38} & 63.79 & 75.86 & \textbf{91.23} & & & \\
        & Pass@3 & 76.72 & 84.48 & \textbf{94.78} & 79.31 & 89.66 & \textbf{96.55} & 74.14 & 79.31 & \textbf{92.98} & & & \\
        \midrule
        \multirow{4}{*}{GPT-5-mini} & Success Rate & 56.32 & 66.97 & \textbf{82.18} & 60.34 & 73.37 & \textbf{89.08} & 52.30 & 60.37 & \textbf{75.29} & 382 & 523 & 453 \\
        \cmidrule(lr){2-14}
        & Pass@1 & 61.21 & 62.28 & \textbf{83.62} & 67.24 & 65.52 & \textbf{91.38} & 55.17 & 58.93 & \textbf{75.86} & & & \\
        & Pass@2 & 73.28 & 78.95 & \textbf{93.97} & 77.59 & 87.93 & \textbf{100.00} & 68.97 & 69.64 & \textbf{87.93} & & & \\
        & Pass@3 & 76.72 & 85.09 & \textbf{97.41} & 81.03 & 94.83 & \textbf{100.00} & 72.41 & 75.00 & \textbf{94.83} & & & \\
        \midrule
        \multirow{4}{*}{GPT-4.1-mini} & Success Rate & 54.08 & 46.52 & \textbf{72.78} & 64.02 & 54.76 & \textbf{76.40} & 44.31 & 37.16 & \textbf{69.03} & 128 & 261 & 190 \\
        \cmidrule(lr){2-14}
        & Pass@1 & 53.45 & 40.71 & \textbf{70.69} & 65.52 & 56.90 & \textbf{74.14} & 41.38 & 23.64 & \textbf{67.24} & & & \\
        & Pass@2 & 62.93 & 59.29 & \textbf{85.34} & 68.97 & 75.86 & \textbf{89.66} & 56.90 & 41.82 & \textbf{81.03} & & & \\
        & Pass@3 & 68.97 & 70.80 & \textbf{89.66} & 74.14 & 86.21 & \textbf{93.10} & 63.79 & 54.55 & \textbf{86.21} & & & \\
        \midrule
        \multirow{4}{*}{\shortstack{Qwen3-Coder\\-30B-A3B\\-Instruct-FP8}} & Success Rate & 54.07 & 30.10 & \textbf{64.38} & 60.83 & 37.25 & \textbf{72.14} & 47.62 & 23.08 & \textbf{57.24} & 590 & 666 & 599 \\
        \cmidrule(lr){2-14}
        & Pass@1 & 45.54 & 32.17 & \textbf{65.22} & 52.73 & 40.35 & \textbf{70.69} & 38.60 & 24.14 & \textbf{59.65} & & & \\
        & Pass@2 & 66.07 & 44.35 & \textbf{76.52} & 72.73 & 52.63 & \textbf{81.03} & 59.65 & 36.21 & \textbf{71.93} & & & \\
        & Pass@3 & 72.32 & 46.09 & \textbf{80.00} & 78.18 & 54.39 & \textbf{82.76} & 66.67 & 37.93 & \textbf{77.19} & & & \\
        \midrule
        \multirow{4}{*}{GPT-4.1} & Success Rate & 24.71 & \textbf{72.06} & 62.82 & 27.01 & \textbf{80.70} & 71.68 & 22.41 & \textbf{63.31} & 54.02 & 39 & 153 & 187 \\
        \cmidrule(lr){2-14}
        & Pass@1 & 26.72 & \textbf{68.97} & 63.79 & 32.76 & \textbf{77.59} & 67.24 & 20.69 & \textbf{60.34} & 60.34 & & & \\
        & Pass@2 & 31.03 & \textbf{79.31} & 74.14 & 36.21 & \textbf{84.48} & 79.31 & 25.86 & \textbf{74.14} & 68.97 & & & \\
        & Pass@3 & 32.76 & \textbf{83.62} & 81.03 & 37.93 & \textbf{89.66} & 87.93 & 27.59 & \textbf{77.59} & 74.14 & & & \\
        \midrule
        \multirow{4}{*}{GPT-5-nano} & Success Rate & 35.34 & 42.06 & \textbf{57.27} & 40.80 & 43.31 & \textbf{60.82} & 29.89 & 40.80 & \textbf{53.76} & 230 & 383 & 394 \\
        \cmidrule(lr){2-14}
        & Pass@1 & 34.48 & 38.37 & \textbf{62.93} & 39.66 & 39.53 & \textbf{63.79} & 29.31 & 37.21 & \textbf{62.07} & & & \\
        & Pass@2 & 44.83 & 50.00 & \textbf{75.00} & 50.00 & 53.49 & \textbf{77.59} & 39.66 & 46.51 & \textbf{72.41} & & & \\
        & Pass@3 & 54.31 & 54.65 & \textbf{81.90} & 56.90 & 60.47 & \textbf{84.48} & 51.72 & 48.84 & \textbf{79.31} & & & \\
        \midrule
        \multirow{4}{*}{GPT-4.1-nano} & Success Rate & 14.24 & 12.89 & \textbf{19.92} & 15.48 & 15.00 & \textbf{24.60} & 13.04 & 11.03 & \textbf{15.38} & 71 & 146 & 178 \\
        \cmidrule(lr){2-14}
        & Pass@1 & 13.79 & 8.93 & \textbf{22.81} & 13.79 & 9.26 & \textbf{31.03} & 13.79 & 8.62 & \textbf{14.29} & & & \\
        & Pass@2 & 19.83 & 16.96 & \textbf{30.70} & 18.97 & 18.52 & \textbf{37.93} & 20.69 & 15.52 & \textbf{23.21} & & & \\
        & Pass@3 & 23.28 & 17.86 & \textbf{32.46} & 25.86 & 20.37 & \textbf{39.66} & 20.69 & 15.52 & \textbf{25.00} & & & \\
        \bottomrule
      \end{tabular}
    }
    \caption{\textbf{Ablation study: impact of meta-skills on DeepSolver.} NSR (No Self-Reflection) and NCL (No Continuous Learning) represent configurations with the respective components disabled. Success Rate represents the overall accuracy across all attempts, while Pass@k (k=1, 2, 3) indicates the proportion of questions with at least one correct answer among the first k attempts. Results are shown for all questions, as well as separately for 0-Level and 1-Level questions. Average time denotes the mean completion time per question in seconds. Bold values indicate the best performance for each model across the three configurations (NSR, NCL, DeepSolver).}
    \label{ablation}
  \end{table*}

\subsection{Demonstrations}
\label{si_demos}

Demonstration examples can be found at \url{https://github.com/CederGroupHub/CASCADE/tree/main/benchmark_tasks_and_results/demonstration/free_form_output}, including task files (with user queries), reference solution code, agent-generated answers, and associated figures.

\clearpage                                   

\bibliography{refs}

\end{document}